\documentclass{article} % For LaTeX2e
\usepackage{iclr2027_conference,times}

\usepackage{amsmath,amsfonts,bm}

\def\eqref#1{equation~\ref{#1}}
\def\1{\bm{1}}

\DeclareMathAlphabet{\mathsfit}{\encodingdefault}{\sfdefault}{m}{sl}
\SetMathAlphabet{\mathsfit}{bold}{\encodingdefault}{\sfdefault}{bx}{n}

\usepackage{xurl}
\usepackage[hidelinks]{hyperref}
\usepackage{url}
\usepackage{graphicx}
\usepackage{booktabs}
\usepackage{amsmath}
\usepackage{enumitem}
\usepackage{makecell}
\usepackage{subcaption}
\usepackage{placeins}

\title{Image Fidelity is Not Field Fidelity:  \\ Joint Thermodynamic Reconstruction and \\ 
Error Localization in Neural Tomography}

\author{Alan Hsu \\
Harvard University\\
Cambridge, MA, USA \\
\texttt{alan.hsu@cfa.harvard.edu} \\
\And
Jenna Samra \\
Center for Astrophysics \textbar\; Harvard \& Smithsonian \\
Cambridge, MA, USA \\
\texttt{jsamra@cfa.harvard.edu} \\
\And
Alin Razvan Paraschiv \\
National Solar Observatory \\
Boulder, CO, USA \\
\texttt{arparaschiv@nso.edu} \\
\And
Liam Connor \\
Harvard University\\
Center for Astrophysics \textbar\; Harvard \& Smithsonian \\
Cambridge, MA, USA \\
\texttt{liam.connor@cfa.harvard.edu} \\
}

\iclrfinalcopy

\begin{document}

\maketitle
\lhead{}

\begin{abstract}
    Neural fields for scientific tomography are optimized from 2D images, but the actual quantity of interest is often a latent 3D physical field. Because the forward map is many-to-one, low 2D image error need not certify a correct 3D field. Moreover, the latent field is not directly supervised during training, and its error cannot be evaluated against truth at deployment. We develop CoroNeRF to jointly optimize 3D electron density and temperature fields directly from multiview, multiline intensities through a differentiable atomic-emission renderer. Using solar coronal tomography as a controlled testbed, we evaluate physical-field recovery and test whether cross-seed instability provides a ground-truth-free-at-inference indicator of local physical-field error. We underscore the following two observations. (i) Image fidelity is not field fidelity: spectral ablations show that limited-channel reconstructions can fit their available observations well while recovering substantially worse fields, whereas evaluation on a common richer probe exposes the discrepancy. (ii) Cross-seed instability ranks local physical-field error across tested matched-model conditions, supported by sparsification and physical signal-strength controls. Seed-deviation projections provide complementary directional validation, but shared forward-model mismatch can still produce incorrect cross-seed consensus. These results characterize joint thermodynamic recovery and the usefulness and limits of seed-based error localization in a controlled, single-scene solar tomography testbed.
\end{abstract}

\section{Introduction} \label{sec:intro}
%%%%%%%%%%%%%%%%%%%%%%%%%%%%%%%%%%%%

Scientific tomographic imaging problems aim to recover latent 3D physical fields where the data available for supervised learning are indirect, multiview 2D measurements, such as projections, line-of-sight (LOS) integrals, or other nonlinear measurements. Neural fields \citep{Mildenhall_2020} and NeRF-style frameworks \citep{Gao_2026} have been developed primarily for 3D scene representation and are a promising paradigm for scientific tomography. These frameworks represent a spatial field as a continuous coordinate-conditioned function and optimize it through a differentiable forward model. Under these frameworks, neural rendering problems are often evaluated as novel view synthesis problems, where success is determined by how well the model reproduces the training images and novel views. However, in scientific inverse problems, notably tomographic reconstruction, such a criterion is incomplete: a model can achieve low 2D image-space error while recovering an incorrect latent 3D physical field, especially when the forward map is non-injective and ill-conditioned, allowing substantially different fields to produce similar measurements. Moreover, in standard novel view synthesis pipelines, error in a physically interpreted latent field is typically not part of the task definition, let alone the training objective. \textbf{Thus, the latent field is not directly supervised during training and its error is unverifiable against truth during deployment,} leaving image agreement as a common but unreliable validation signal.

We investigate this limitation through joint reconstruction of the solar corona's 3D electron density and temperature. The solar corona is the outermost layer of the Sun, composed of hot plasma that is optically thin across a large wavelength range. Multiline 2D intensity observations are LOS integrals of 3D plasma emissivities (analogous to the RGB$\sigma$ radiance proxies in standard NeRFs), and these emissivities are functions of the latent 3D plasma density and temperature fields. The goal is therefore not merely to reconstruct each line's emitting structure, but to recover a shared density-temperature state that explains all the observed channels. This inverse problem is highly ill-posed as distinct 3D density-temperature fields can produce nearly indistinguishable measurements. There is a standard geometric LOS degeneracy: since each pixel is an integral along a ray, a redistribution of the emissivities along that ray can still produce the same image. In addition, without sufficient spectral coverage, there is a plasma thermodynamic degeneracy: many density-temperature realizations can produce similar multiline emissivity values because the line-response functions are not jointly injective.

We thus introduce CoroNeRF, a differentiable multiline neural tomography framework for the direct joint recovery of 3D electron density and temperature from multiview spectral line-intensity images. Instead of learning the RGB$\sigma$ radiance proxy of the scene, CoroNeRF parameterizes a neural plasma field that maps 3D position $\mathbf{x}$ to electron density $\log_{10}n_{\mathrm e}(\mathbf{x})$ and temperature $\log_{10}T_{\mathrm e}(\mathbf{x})$. A differentiable line-emission renderer then maps these fields to multichannel spectral line intensities using precomputed emissivity tables on samples along each camera ray. All channels are synthesized from the same density and temperature fields, which are optimized jointly using only image supervision. Within this framework, we examine how observational information affects thermodynamic recovery and whether cross-seed instability can localize the physical-field error without ground truth at inference.

Our contributions are as follows: 
\begin{enumerate}[leftmargin=*]
    \item \textbf{Direct joint thermodynamic neural tomography.} We develop CoroNeRF to optimize 3D electron density and temperature directly from multiview, forbidden-line intensities through a differentiable atomic-emission renderer. At each spatial location, all modeled lines share the same inferred electron density and temperature, coupling the channel reconstructions through their atomic-emission responses.

    \item \textbf{Image fidelity is not field fidelity.} In a controlled synthetic scene, spectral ablations show that limited-channel reconstructions can fit their available observations well while recovering substantially worse physical fields. A common richer probe exposes these discrepancies, while noise-view experiments characterize reconstruction performance under reduced observational information.

    \item \textbf{Empirical field-error localization and its limits.} Cross-seed instability ranks local physical-field error across tested matched-model conditions, with emissivity, radial, sparsification, and seed-stability controls. Seed-deviation projections provide complementary directional validation, while a specified abundance mismatch demonstrates that shared model error can produce incorrect consensus.
\end{enumerate}

\section{Related Work} \label{sec:related}
%%%%%%%%%%%%%%%%%%%%%%%%%%%%%%%%%%%%

In the past two decades, various tomography methods have been used for coronal reconstruction, such as recovering coronal density from LASCO-C2 white-light images via classical regularized-least-squares \citep{Frazin_2002} and 3D coronal density and temperature through EUV tomography \citep{Frazin_2009}. Vector coronal tomographic inversion theory has also been developed for magnetic field inversion from spectropolarimetric observations, validated on both simulations \citep{Kramar_2006, Kramar_2007, Kramar_2013, Kramar_2026} and real observations \citep{Kramar_2016}. These approaches use classical inverse-problem formulations with explicit regularization or prior structure, rather than learned neural representations coupled with an end-to-end differentiable nonlinear emission model.

Implicit neural representations and related learned inverse models have increasingly been used for scientific tomography. In medical imaging \citep{Wang_2024, Kabika_2026}, uncertainty-aware null-space networks have been studied for MRI \citep{Angermann_2023}. Neural reconstruction has also been used for heterogeneous macromolecular structures \citep{Zhong_2021}. Related neural-field approaches have been studied for physical flow field tomography \citep{Molnar_2022}, seismic full-waveform inversion \citep{Sun_2023}, black hole dynamic tomography \citep{Levis_2022, Feng_2026}, and thermal tomography in inverse heat conduction problems \citep{Zhong_2026}. NeRFs have also been used for solar coronal reconstruction: \cite{Ramos_2023} reconstructs density from polarized-brightness observations, SuNeRF \citep{Jarolim_2024} models wavelength-specific EUV emission and absorption, while SuNeRF-CME \citep{Jarolim_2026} reconstructs time-dependent electron density through Thomson-scattering observations. Related efforts have reported density-temperature reconstruction from multithermal EUV imaging \citep{swri_2024, Jarolim_2025}. To our knowledge, CoroNeRF is the first neural-field framework to directly and jointly reconstruct 3D coronal electron density and temperature from multiview forbidden infrared-line intensities through a differentiable atomic-emission model.

Methods for uncertainty quantification in NeRFs have also been developed over the past five years. Density-aware NeRF ensembles employ deep ensembles to identify regions unobserved during training \citep{Sunderhauf_2023}; full probabilistic frameworks such as Stochastic NeRFs \citep{Shen_2021} and Conditional-Flow NeRFs \citep{Shen_2022} quantify uncertainty by learning the distribution over NeRFs. Prior post-hoc methods also study uncertainty in conventional radiance fields: Bayes' Rays \citep{Goli_2024} evaluates geometric depth uncertainty, while FisherRF \citep{Jiang_2024} uses Fisher information. Finally, concurrent work on sparse-view CT shows that strong global uncertainty-error association can deteriorate substantially within the reconstructed object, and that error shared across independently trained members can remain invisible to ensemble spread \citep{Zhao_2026}. We study a complementary setting involving coupled density-temperature recovery through nonlinear multiline atomic-emission responses, characterizing when cross-seed instability localizes physical-field error and when shared forward-model mismatch causes it to fail.
\section{Methodology: Differentiable Multiline Neural Tomography} \label{sec:methods}
%%%%%%%%%%%%%%%%%%%%%%%%%%%%%%%%%%%%

CoroNeRF parameterizes a neural plasma field $f_\theta(\mathbf{x}) \mapsto (\log_{10} \hat{n}_{\mathrm{e}, \theta}(\mathbf{x}), \log_{10}\hat{T}_{\mathrm{e},\theta}(\mathbf{x}))$. Rather than first inverting observed line ratios, it jointly fits the absolute intensities of all lines through their emissivity responses. Density and temperature diagnostic information therefore enters implicitly through the forward model, while multiview geometry constrains the 3D distribution.

The observations are multiview, multichannel line-intensity images. Let $v \in \mathcal{V}$ index viewpoints, $p \in \Omega_v$ index valid image pixels under the view mask, and $c \in \{1, \ldots, C\}$ index the spectral line channels. We write $q = (v, p)$ for the ray associated with pixel $p$ in view $v$. Each ray is then parameterized as $\mathbf{x}_q(t) = \mathbf{o}_q + t \mathbf{d}_q, \; t \in [t_s(q), t_f(q)]$, where $\mathbf{o}_q$ is the ray origin and $\mathbf{d}_q$ is the unit direction. The observed multichannel intensity vector for this ray is $\mathbf{y}_q = (y_{q,1}, \ldots, y_{q,C}) \in \mathbb{R}^{C}$. The training ray dataset is then given by $\mathcal{D}_{\mathrm{ray}} = \left\{ \left(\mathbf{x}_q(\cdot), \mathbf{y}_q\right): q = (v,p), \; v \in \mathcal{V}_{\mathrm{train}}, \; p \in \Omega_v \right\}$.

Given a neural plasma field, the differentiable line-emission renderer maps physical variables to predicted line intensities. For channel $c$, the renderer first evaluates a channel-dependent emissivity $\epsilon_c(n_{\mathrm e}, T_{\mathrm e}, r)$, where $r$ is the heliocentric radius, using a precomputed atomic-physics lookup table from the CHIANTI database (\cite{Dere_1997, Dere_2023}, Appendix \ref{app:emissivity_table}). It then integrates the emissivities along each ray to produce the predicted line intensity: 

\begin{equation}
    \hat{I}_{\theta, q, c} = \int_{t_s(q)}^{t_f(q)} \epsilon_c \! \left(\hat{n}_{\mathrm{e}, \theta}(\mathbf{x}_q(t)), \hat{T}_{\mathrm{e}, \theta}(\mathbf{x}_q(t)), r(\mathbf{x}_q(t)) \right) \, dt.
    \label{eq:los_eps_integral}
\end{equation}

In practice, we compute LOS integrals using numerical quadrature at the fixed sampling resolution described in Appendix \ref{app:sampling_quadrature}. The inverse problem is then latent physical-field recovery of $n_{\mathrm{e}}(\mathbf{x})$ and $T_{\mathrm{e}}(\mathbf{x})$: we optimize the neural plasma field parameters $\theta$ using image supervision only:

\begin{equation}
    \theta^\star = \arg\min_\theta \frac{1}{|\mathcal{D}_{\mathrm{ray}}|} \sum_{(x_q(\cdot), y_q) \in \mathcal{D}_{\mathrm{ray}}} \frac{1}{C} \sum_{c = 1}^{C} \rho(\hat{I}_{\theta, q, c}, y_{q, c}).
\end{equation}

For most experiments, we use a fixed-scale asinh image loss $\rho(\hat{I}, y) = | \operatorname{asinh}(\hat{I}/s_c) - \operatorname{asinh}(y/s_c) |$, where $s_c$ is a channel-dependent fixed characteristic intensity scale, chosen during development so as to not allow bright channels to dominate. While our images have heteroscedastic Gaussian noise, image-space fidelity is not the ultimate objective: we thus use the asinh-L1 image loss to compress the intensity dynamic range. Among the tested objectives, it yields lower field error than the inverse-variance-weighted squared-residual objective (Appendix \ref{app:loss_ablation}).

\section{Experimental Setup and Evaluation} \label{sec:setup}
%%%%%%%%%%%%%%%%%%%%%%%%%%%%%%%%%%%%

\begin{figure}[!th]
  \centering
  \newcommand{\wLeft}{0.37}
  \newcommand{\wRight}{0.60}
  \begin{subfigure}[c]{\wLeft\textwidth}
    \centering
    \includegraphics[width=\linewidth]{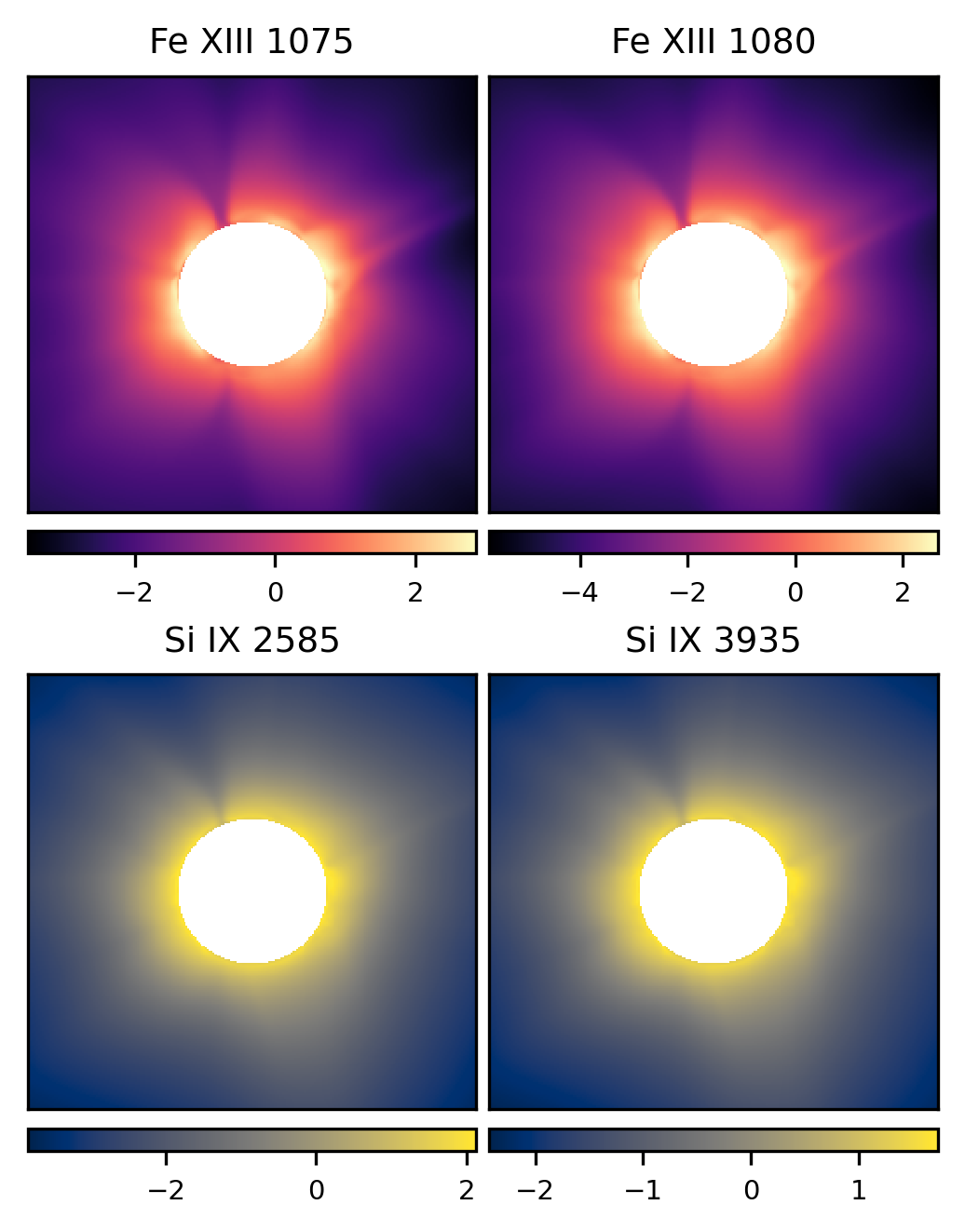}
  \end{subfigure}
  \hfill
  \begin{subfigure}[c]{\wRight\textwidth}
    \centering
    \includegraphics[width=\linewidth]{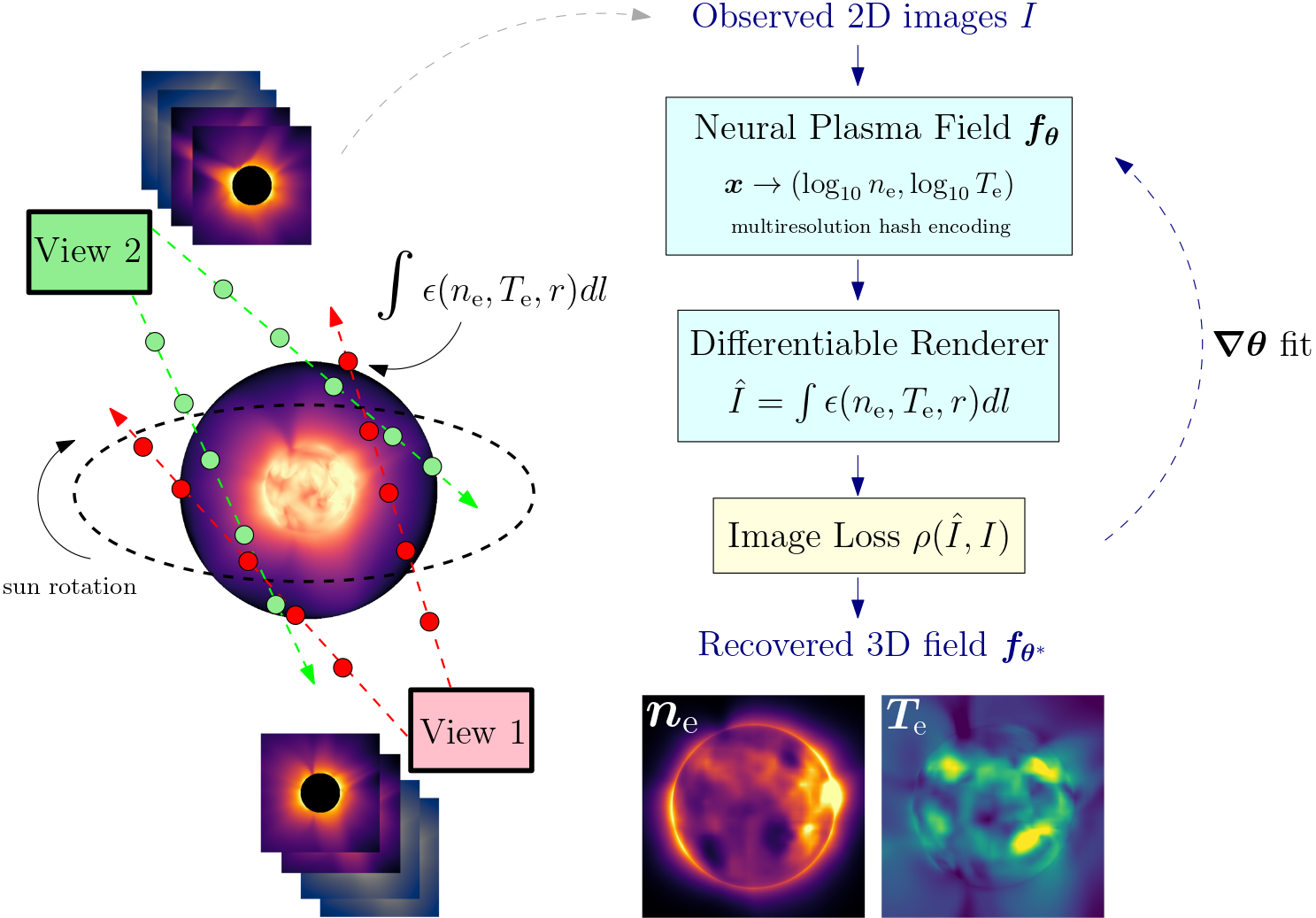}
  \end{subfigure}
  \vspace{-9pt}
  \caption{CoroNeRF overview. The left $2\times2$ grid displays multiline coronal observations of four spectral channels (Fe XIII $1075/1080$ nm, Si IX $2585/3935$ nm) forward-rendered from a PSI cube. The observations are used to fit a multiresolution hash-grid neural plasma field through the same differentiable renderer by minimizing an image loss (right panel). The fitted neural field yields jointly recovered 3D density and temperature fields.}
  \label{fig:overview}
\end{figure}

We construct synthetic observations (left panel of Figure \ref{fig:overview}) of four spectral channels (Fe XIII $1075/1080$ nm, Si IX $2585/3935$ nm) from one ground truth PSI MAS (MHD) coronal cube \citep{Mikic_2007, Lionello_2008} based on Carrington Rotation 2283 (April-May 2024). The GT field has ($299$, $142$, $154$) longitude-colatitude-radius voxels spanning the full sphere and ranging from $[1,30]$ R$_\odot$. The master dataset consists of 3000 views evenly spaced in longitude at $0^\circ$ latitude: we mimic observational coverage expected from real Earth measurements, but the viewing geometry is a static-rotational tomography idealization. Each view is 256 by 256 pixels with a FOV of $[-3, +3]$ R$_\odot$, and each pixel corresponds to a particular ray through the coronal field. Because the corona is optically thin and faint ($\approx10^{-6}$ of the central photospheric disk brightness), it is often observed by occulting the bright solar disk, as done in a coronagraph. We mirror this procedure by masking pixels within $1$ R$_\odot$, which additionally confines the LOS integral to the $r \ge 1$ R$_\odot$ region where the emissivity model is well-defined. Consequently, all observation-space training and evaluation images apply the occulting mask. The canonical training set contains $300$ evenly-spaced views ($19.66$M rays, $78.64$M channel measurements, before the occulting mask), excluding a $30^\circ$ evaluation arc and a $3^\circ$ guard region around it: evaluation uses the arc, while the guard region is excluded from both sets. We chose this holdout protocol because it explicitly leaves a section of the corona views unseen by the model, which is a harder interpolation test rather than an interleaved-view held-out set.

We summarize each observational condition, or experiment family, by the tuple $(\eta, V, \Lambda, a)$. We add heteroscedastic Gaussian noise (Equation \ref{eq:noise_model}) to our images, with the shot-noise variance multiplier $\eta$ (e.g. $\eta=9$ is $\times3 \sigma$). $V$ is the number of views, $\Lambda$ is the set of spectral channel indices, and $a \neq 1$ is a forward-model elemental abundance mismatch scale (Appendix \ref{app:obs_model}). We define the canonical noisy condition as $(\eta = 9, V = 300, \Lambda = \text{all 4}, a = 1 )$, used later in Section \ref{sec:results}.

Our joint-reconstruction model uses a multiresolution hash-grid encoder (Instant NGP-style, \cite{Muller_2022}), with separate decoder heads for density and temperature predictions (Appendix \ref{app:hash_grid}). We use the AdamW optimizer for $60$k steps with batches of $1024$ rays, optimized using a fixed-scale asinh loss (Appendices \ref{app:loss_ablation}, \ref{app:compute_ablation}). We evaluate performance using field-space mean absolute error on the inner coronal radial band $[1.1,2.0]$ R$_\odot$, as well as held-out image-space loss (Appendix \ref{app:metrics}). We show a visualization of the renderer and the training process in the right panel of Figure \ref{fig:overview}. Training a single reconstruction takes $\approx3$ GPU-hours on an NVIDIA A$100$, and rendering a 300-view dataset takes $\approx10$-$100$ minutes on an NVIDIA RTX 4090 depending on the resolution and integration step size (Appendix \ref{app:compute_ablation}). 

We stress that this is a controlled study using one static coronal scene. Thus, we are evaluating joint thermodynamic reconstruction and within-scene error localization, not a probabilistic calibration or generalization across multiple coronal states. The objective family and architecture were selected based on physical-field recovery on this scene; for noise-level experiments, the asinh scale $s_c$ is fixed at its baseline for $\eta \le 1$ and scaled by $\sqrt{\eta}$ for $\eta > 1$ (Appendix \ref{app:noise_view_bench}). All synthetic generation and inversion experiments also use the same discretized renderer (Appendix \ref{app:sampling_quadrature}), so the study evaluates reconstruction under a matched forward model except for abundance perturbations described in Appendix \ref{app:obs_model}. 
\section{Results} \label{sec:results}
%%%%%%%%%%%%%%%%%%%%%%%%%%%%%%%%%%%%

\subsection{Joint Thermodynamic Reconstruction and Measurement Limitations}
\label{subsec:baseline}

\begin{figure}[!th]
	\begin{center}
		\includegraphics[width=1\linewidth]{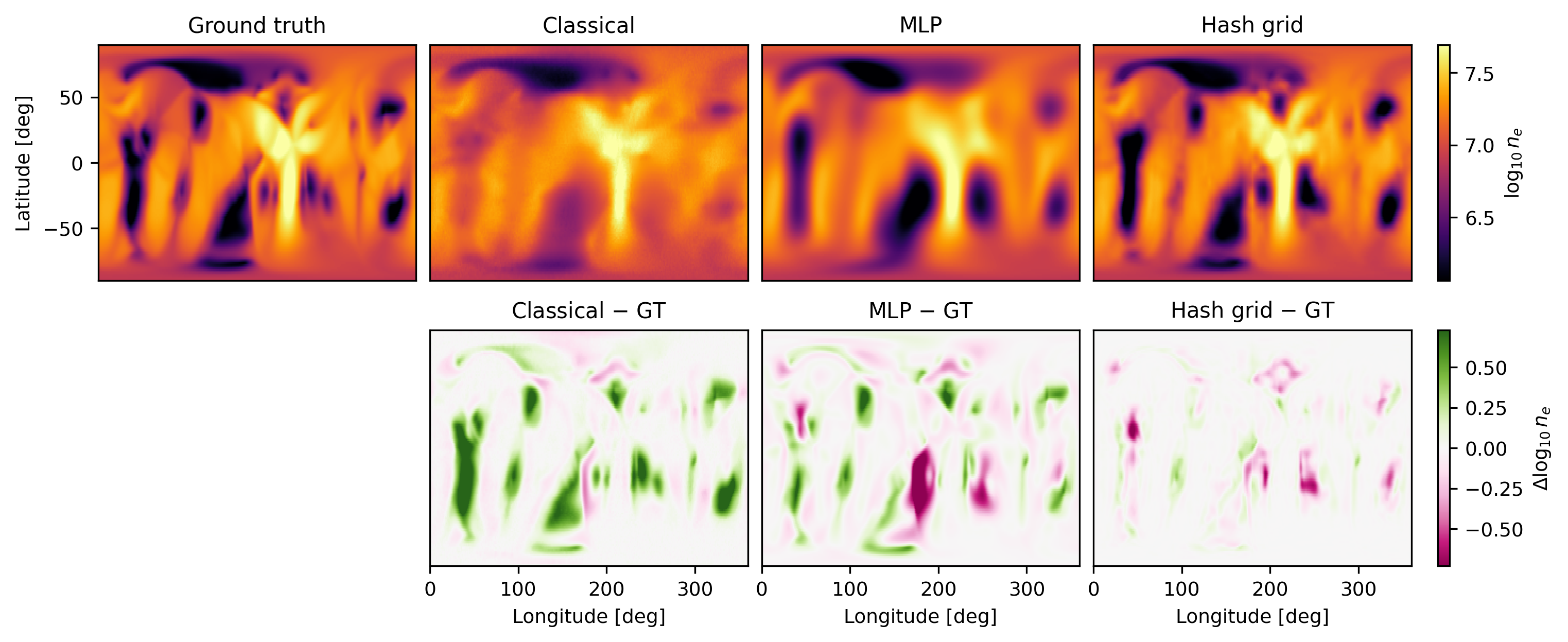}
	\end{center}
    \vspace{-9pt}
	\caption{Density-field reconstruction (given GT temperature) at $1.51$ R$_\odot$. Each panel is a shell unraveled on a latitude-longitude plot. The top row shows the ground truth and reconstructions using a classical grid with Tikhonov regularization, positional encoder + MLP, and hash grid + decoder MLP representations. The bottom row shows corresponding residuals $\Delta \log_{10} n_e = \log_{10} n_{e, \mathrm{pred}} - \log_{10} n_{e, \mathrm{GT}}$. The hash grid is able to reproduce sharper small-scale features, substantially reducing inner-band $\operatorname{MAE}$ relative to the MLP by more than $2\times$ and classical grid methods by about $4\times$.}
	\label{fig:benchA_reconstruction}
\end{figure}

We first isolate the spatial representation by reconstructing density while supplying GT temperature and then select the hash-grid representation for the joint density-temperature inversion. Figure \ref{fig:benchA_reconstruction} shows density-field reconstruction results at $1.51$ R$_\odot$, trained using two matched-model, noiseless intensity channels (Fe XIII $1075/1080$) while given the ground truth temperature field. Latitude-longitude plots (flattened spherical shells) of the reconstructed density are shown on the top, while the signed residuals are shown on the bottom. We use $\operatorname{MAE}_{\text{inner}}(\log_{10}n_{\rm e})$, the mean absolute error of log-density in the inner radial band $[1.1, 2.0]$ R$_\odot$, as our main metric. Under the tested grid resolution, architectures, and optimization settings, the multiresolution hash grid + decoder head attains the lowest reconstruction error, performing over twice as well ($0.042$ dex $\operatorname{MAE}$) as a standard positional encoder + MLP ($0.097$ dex $\operatorname{MAE}$), and about four times as well as a classical voxelized representation + Tikhonov regularizer ($0.168$ dex $\operatorname{MAE}$). These comparisons are intended to isolate the effect of spatial representation under a common emission renderer, rather than to benchmark complete solar-tomography systems, which are generally tailored to different observations and inversion settings. See Appendix \ref{app:representation_baselines} for detailed results.

\begin{figure}[!th]
	\begin{center}
		\includegraphics[width=1\linewidth]{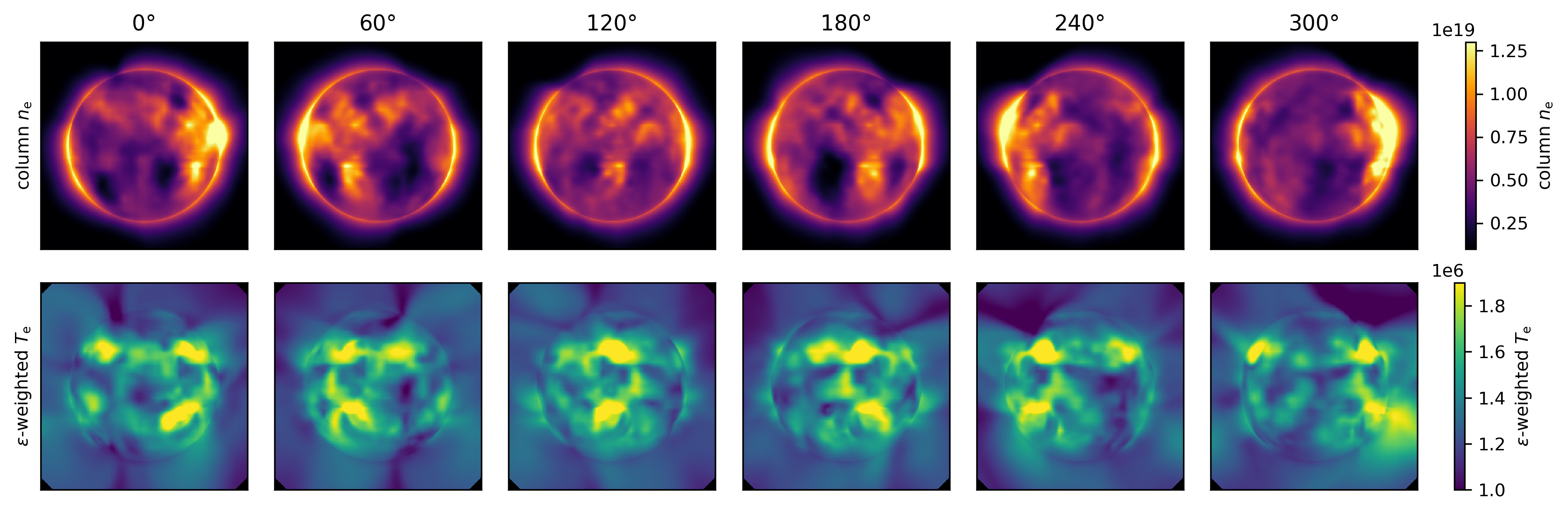}
	\end{center}
    \vspace{-9pt}
	\caption{LOS projections of the jointly-reconstructed column density (top) and total-emissivity-weighted temperature (bottom), using a multiresolution hash-grid representation trained on all four spectral channels (Appendix \ref{app:los_viz}).}
	\label{fig:benchB_filmstrip}
\end{figure}

Next, we jointly reconstruct density and temperature using all four spectral channels. The resulting inner-band $\operatorname{MAE}$s are $0.039$ dex for density ($9.2\%$ AbsRel) and $0.014$ dex for temperature ($3.2\%$ AbsRel). Figure \ref{fig:benchB_filmstrip} visualizes the recovered fields through column-density and emissivity-weighted-temperature projections, and Figure \ref{fig:benchB_multi_radius_shell} in Appendix \ref{app:spectral_bench} compares reconstructed fields with ground truth at three radii. We find empirical evidence that increasing spectral coverage substantially reduces the density-temperature degeneracies: single-line models are underconstrained because the observed intensity depends jointly on density and temperature through a single emissivity response, allowing multiple density-temperature states to produce similar measurements. Because each line also has a finite temperature-response range, additional lines with different thermodynamic responses provide complementary constraints that substantially reduce these degeneracies: a pair of lines drops the $n_{\rm e}$ $\operatorname{MAE}$ by a factor of $5$-$10$, and using four lines further halves the error. Finally, additional spectral lines also reduce the radial $\operatorname{MAE}$ at larger radii (Figure \ref{fig:benchB_radial_curves}). 

\begin{figure}[!th]
	\begin{center}
		\includegraphics[width=1\linewidth]{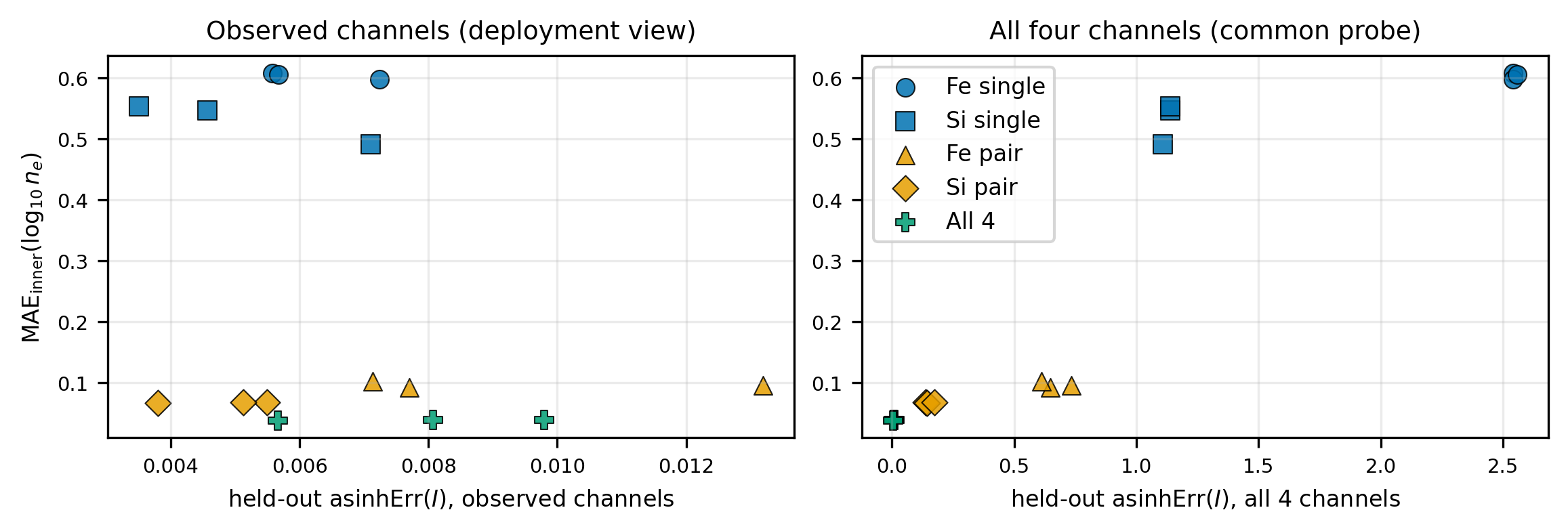}
	\end{center}
    \vspace{-9pt}
	\caption{Image fidelity is not field fidelity. We show 3D density field-space $\operatorname{MAE}$ vs 2D held-out image-space loss. Each marker is a trained model (color $=$ number of spectral lines, shape $=$ the specific line set). The left plot uses held-out image error measured on each model's own channels, which is a GT-free image-space validation quantity available on held-out observed views. Single-channel models achieve lower observed-channel image error than the four-line model despite having substantially larger field error. Evaluation on a common four-line probe (right panel) exposes these failures and largely restores the density-error ranking, showing that the low observed-channel image error of these configurations reflects limited spectral coverage rather than correct field recovery.}
	\label{fig:benchB_scatter}
\end{figure}

We further show that \textbf{image fidelity is not field fidelity}: a low image-space error need not certify a low field-space error. Figure \ref{fig:benchB_scatter} shows a per-model scatter of field-space $\operatorname{MAE}$ vs 2D held-out validation image error (color $=$ number of spectral lines, shape $=$ the specific line set). The left panel uses the held-out image error measured on each model's own channels, which is a GT-free image-space validation quantity available on held-out observed views. We see that single-channel experiments fit their observed images better than the four-line models, yet they recover a far worse density field. The right panel recomputes the image-space evaluation on a common four-channel set across all models with the same fixed observational setup, showing that own-channel image error is not comparable as a certificate of field fidelity across spectral configurations. In the density comparison, the common probe largely restores the field-error ranking. Temperature reconstructions likewise exhibit the same discrepancy, although the common-probe image error does not perfectly order temperature error across line sets (Figure \ref{fig:benchB_scatter_appendix}).

\begin{figure}[!th]
	\begin{center}
		\includegraphics[width=1\linewidth]{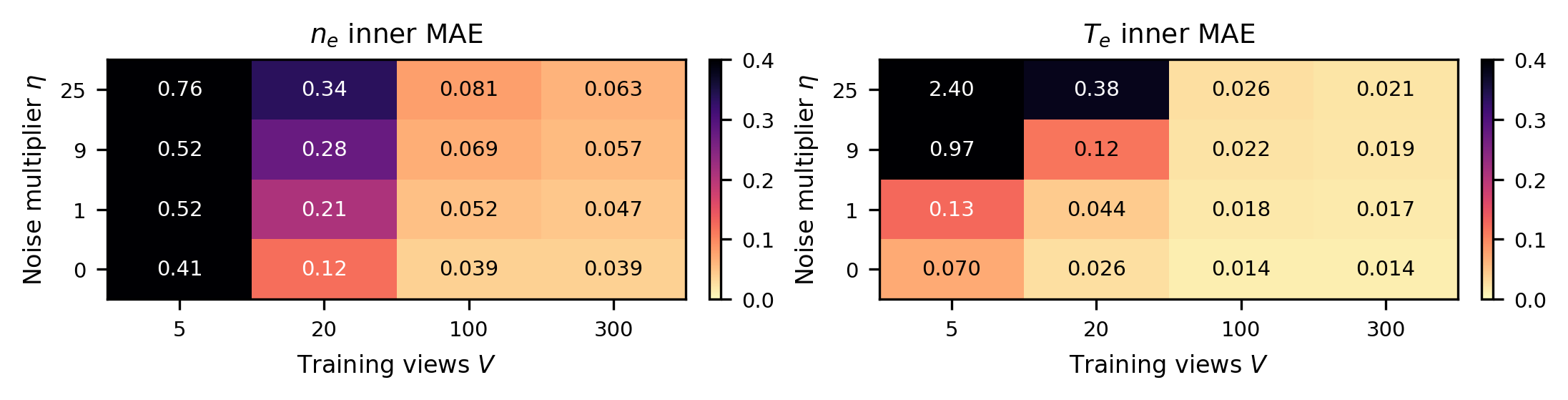} 
	\end{center} 
    \vspace{-9pt}
	\caption{Noise-view sweep, fixed at 4 matched-model spectral channels, of $n_{\rm e}$ and $T_{\rm e}$ $\operatorname{MAE}$. With at least $100$ views, field recovery remains accurate across the tested noise levels under the noise-matched loss. Severe angular undersampling substantially increases error and makes the reconstruction much more vulnerable to noise.} 
	\label{fig:benchC_heatmaps} 
\end{figure}

Finally, we assess the robustness under sparse-view and high-noise regimes. Figure \ref{fig:benchC_heatmaps} crosses four noise levels, $\eta \in \{0, 1, 9, 25\}$, with four view counts, $V \in \{5, 20, 100, 300\}$. For $V \le 20$, density $\operatorname{MAE}$ increases several-fold relative to $V \ge 100$, and severe angular undersampling substantially increases sensitivity to noise. However, for $V \ge 100$, increasing the noise scale from $1 \rightarrow 25$ only modestly increases the $\operatorname{MAE}$ (for example, $0.052 \rightarrow 0.081$ at $100$ views). We conclude that within this scene, reconstructions with at least $100$ views retain low field error across the tested noise levels. We additionally note that because $s_c$ sets the scale of the asinh training loss, we scale it with the measurement-noise standard deviation; Figure \ref{fig:benchC_heatmaps} therefore compares reconstructions under a noise-matched loss.

\subsection{Cross-Seed Instability as a GT-free-at-Inference Error-Ranking Proxy}
\label{subsec:ensemble_disagreement}

\begin{figure}[!th]
	\begin{center}
		\includegraphics[width=1\linewidth]{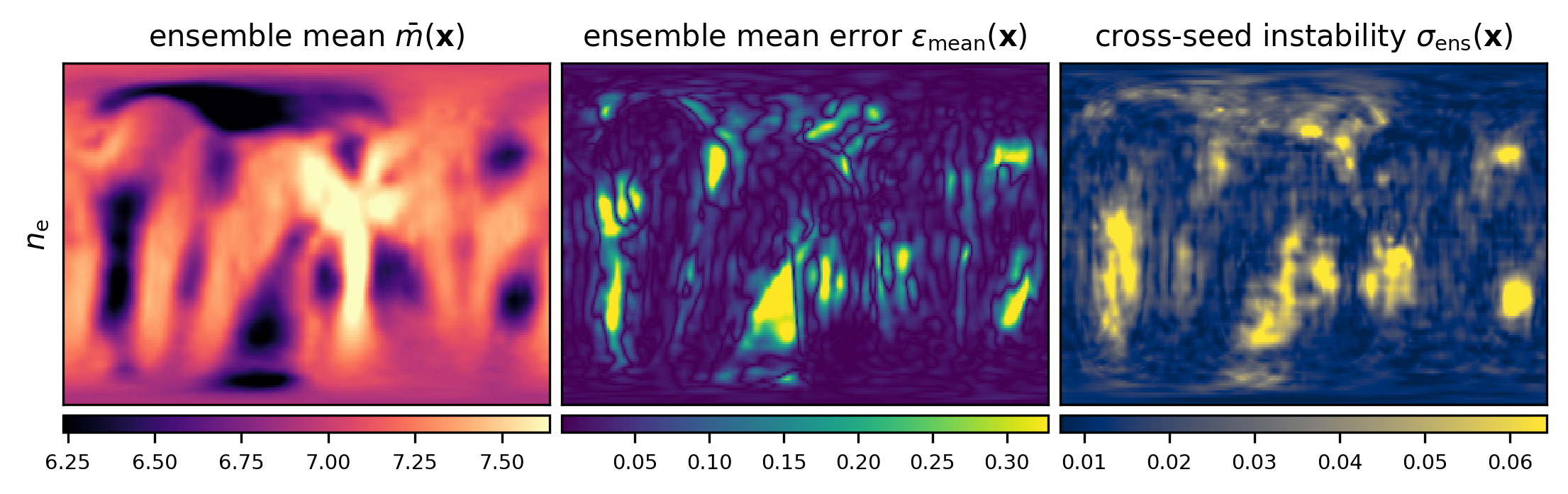}
	\end{center}
    \vspace{-9pt}
	\caption{Ensemble panels of the 3D density field at $1.51$R$_\odot$ using the canonical high-noise condition. We show the ensemble mean $\overline{m}(\mathbf{x})$, ensemble-mean error $\epsilon_{\rm mean}(\mathbf{x})$, and cross-seed instability $\sigma_{\rm ens}(\mathbf{x})$. The maps exhibit similar spatial structure, which we quantify using correlations below.}
	\label{fig:benchF_seed_disagreement}
\end{figure}

We evaluate whether standard seed ensembles provide useful local error rankings for the jointly reconstructed thermodynamic fields. We train $K = 10$ models for each condition, varying only the training seeds \citep{Lakshminarayanan_2017} while holding the model architecture, observational condition, and dataset fixed. The per-voxel standard deviation in the field predictions across the models is the cross-seed instability $\sigma_{\rm ens}$ (Appendix \ref{app:cross_seed}). Figure \ref{fig:benchF_seed_disagreement} shows reconstructions at $1.51$ R$_\odot$ with the canonical high-noise regime for ensemble mean $\overline{m}(\mathbf{x})$, absolute ensemble-mean error $\epsilon_{\rm mean}(\mathbf{x})$, and cross-seed instability $\sigma_{\rm ens}(\mathbf{x})$. \textbf{Large $\sigma_{\rm ens}$ identifies regions where the recovered field is sensitive to training seeds under the fixed pipeline, while small $\sigma_{\rm ens}$ identifies regions of cross-seed consensus.} We note that consensus is not a correctness guarantee, because error shared by all ensemble members (such as a common-mode forward-model discrepancy) will go undetected.

\begin{figure}[!th]
	\begin{center}
		\includegraphics[width=1\linewidth]{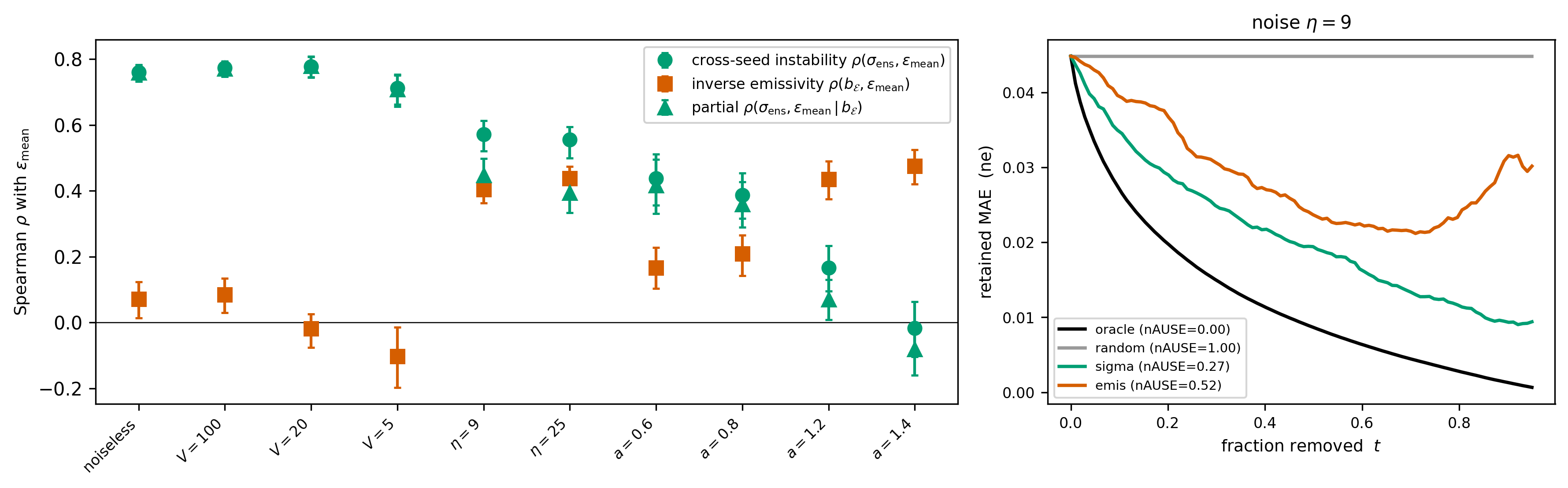}
	\end{center}
    \vspace{-9pt}
	\caption{Left panel plots Spearman rank correlation between the true ensemble mean error $\epsilon_{\rm mean}$ and $\sigma_{\rm ens}$ (green circle), inverse-emissivity $b_{\mathcal{E}}$ (orange), and partial correlation $\rho(\sigma_{\rm ens}, \epsilon_{\rm mean} \, | \, b_{\mathcal{E}})$ (green triangle), showing that the rank association persists after controlling for the oracle inverse-emissivity proxy. Error bars are $95\%$ within-scene longitude-slice bootstrap intervals. We find that $\sigma_{\rm ens}$ localizes error in three matched-model regimes: noiseless, sparse-view, and high-noise, outperforming the emissivity proxy $b_\mathcal{E}$. The partial correlation shows that the rank association remains after conditioning on an oracle inverse-emissivity proxy. Finally, we note that the localization of $\sigma_{\rm ens}$ deteriorates substantially under positive Si over-scaling and approaches chance for density at $a = 1.4$, but under-scaling retains partial localization ability; while $b_\mathcal{E}$ can still track the high-error, low-emissivity regions in this regime, it requires the GT field and is thus used only as an oracle baseline. The right panel consists of sparsification curves for the canonical noisy condition, showing that $\sigma_{\rm ens}$ tracks the oracle more closely than does the baseline $b_\mathcal{E}$ (nAUSE of $0.27$ compared to $0.52$).}
	\label{fig:uq_baseline_ladder}
\end{figure}

We further examine whether rank association is explained by local signal strength. The left panel in Figure \ref{fig:uq_baseline_ladder} shows the Spearman rank spatial correlation for the inner band density field between $\epsilon_{\rm mean}$ and (i) $\sigma_{\rm ens}$ and (ii) an oracle inverse-emissivity signal-strength baseline $b_{\mathcal{E}}$. Error bars are $95\%$ within-scene longitude-slice bootstrap intervals computed on the inner radial band using $2000$ resamples. We find that $\sigma_{\rm ens}$ localizes error across three matched-model regimes: noiseless at $300$ and $100$ views ($0.76$-$0.77$), sparse-view at $20$ and $5$ views ($0.71$-$0.78$), and high-noise ($0.55$-$0.57$); $b_\mathcal{E}$ has weak rank correlation with error in the noiseless and sparse-view regimes, while $\sigma_{\rm ens}$ localizes strongly. Moreover, the partial correlation $\rho(\sigma_{\rm ens}, \epsilon_{\rm mean} \, | \, b_{\mathcal{E}})$ remains positive across matched-model conditions, suggesting that $\sigma_{\rm ens}$ captures error structure beyond what is explained by signal strength alone. This association also remains positive within radial shells (Tables \ref{tab:uq_pershell_correlations_ne} and \ref{tab:uq_pershell_correlations_temp}), indicating that it is not driven solely by radial stratification, although its magnitude varies strongly under five views and high noise. Finally, at the strongest tested Si over-scaling ($a = 1.4$), $\sigma_{\rm ens}$ approaches chance for density, showing that error shared across ensemble members can produce incorrect cross-seed consensus; although $b_\mathcal{E}$ can still localize low-signal, high-error regions in this regime, it requires the GT field and is used only as an oracle diagnostic baseline (Tables \ref{tab:uq_correlations_ne} and \ref{tab:uq_correlations_T}).

Beyond correlation, we use a normalized Area Under the Sparsification Error curve $\operatorname{nAUSE}$ \citep{Ilg_2018, Poggi_2020} to quantify how well a ranking proxy approximates the oracle $\epsilon_{\rm mean}$ ($\operatorname{nAUSE}$ is $0$ for oracle-level sparsification and $1$ for the expected random-reference performance). We find that $\sigma_{\rm ens}$ ($\operatorname{nAUSE}$ $0.11$-$0.30$) is substantially closer to oracle sparsification than the tested baselines. In particular, our inverse-emissivity baseline $b_\mathcal{E}$ is consistently weaker than $\sigma_{\rm ens}$ in the matched-model conditions ($\operatorname{nAUSE}$ $0.47$-$1.25$) and becomes near- or worse-than-random in sparse-view cases. Even in the high-noise regime where $b_\mathcal{E}$ can track the error, $\sigma_{\rm ens}$ performs better ($\operatorname{nAUSE}$ $0.27$ vs $0.52$ for $\times9$, right panel in Figure \ref{fig:uq_baseline_ladder}). Finally, consistent with spatial correlations, under abundance mismatch $\sigma_{\rm ens}$ approaches chance ($\operatorname{nAUSE} = 0.93$ for density at $a = 1.4$). See Appendix \ref{app:sparsification} (Figure \ref{fig:uq_sparsification_curves}) for more details on sparsification analysis and temperature field results.

\begin{figure}[!th]
	\begin{center}
		\includegraphics[width=1\linewidth]{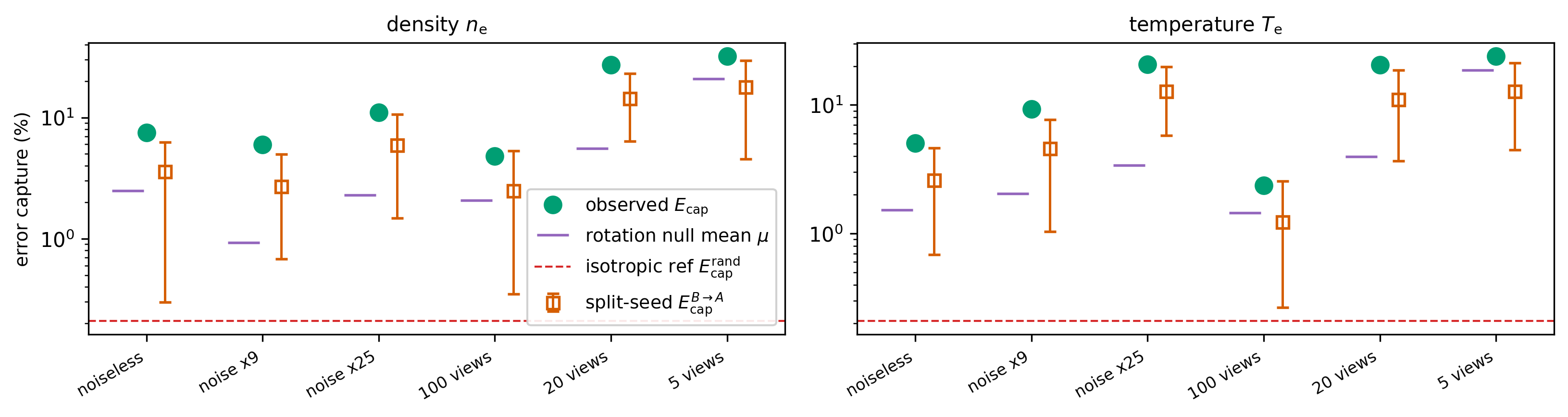}
	\end{center}
    \vspace{-9pt}
	\caption{Per-field error-energy capture computed by projecting the error onto ensemble deviation subspaces. Rank-9 seed-deviation spans capture $2.4$-$32.1\%$ of the signed field error energy (green), which is above the $0.2\%$ isotropic reference (dashed red). Split-seed evaluation, using span and error from disjoint seed halves (orange), shows that the directional overlap persists when the span and target error are constructed from disjoint seed subsets. We additionally find that the original unrotated capture exceeds the longitude-rotation-null mean (purple) in all tested matched-model regimes and lies at the top of the tested rotation distribution in 11 of 12 combinations. Finally, we note that capture is largest under sparse-view and high-noise regimes.}
	\label{fig:uq_errorcap_ladder}
\end{figure}

Next, we provide a complementary field-space consistency test by computing the directional per-field energy overlap $E_{cap}$ (Equation \ref{eq:app_ecap}), or fraction of the squared $L^2$ norm of the signed field error $e$ (whose magnitude is $\epsilon_{\rm mean}$) that lies within the seed-deviation subspace. These percentages are computed on the full $1.05$-$3$ R$_\odot$ control grid, as opposed to the inner radial band used by localization metrics. We also compute the overlap enrichment relative to the rank-matched isotropic expectation $E_{\rm cap}^{\rm rand} = r/M$. Across the three matched-model regimes, shown in Figure \ref{fig:uq_errorcap_ladder}, the observed rank-9 deviation span (green) captures $4.8$-$32.1\%$ of the density signed-error energy and $2.4$-$24.0\%$ of the temperature signed-error energy, strongest under severe view sparsity, and modest under full coverage (all of which are substantially above the isotropic reference of $0.2\%$ in dashed red). The corresponding enrichment ranges between $23$-$154\times$ for density and $11$-$115\times$ for temperature. We additionally note that even under forward model mismatch, $5$-$58\times$ enrichment can persist above the rank-matched isotropic expectation. 

While a high $E_{\rm cap}$ indicates directional alignment, it could be inflated by (i) reusing the same finite ensemble to construct $e$ and the deviation span, or (ii) generic spatial structure. We provide two corresponding controls to test this. First, we split the $K = 10$ ensemble into two $5$-seed groups $A$ and $B$ and compute the error from $B$ captured by the field deviation subspace of $A$ ($E_{\rm cap}^{B \rightarrow A}$), repeated over all $252$ splits. Figure \ref{fig:uq_errorcap_ladder} shows the mean and $2.5$th and $97.5$th percentile interval of the split-sensitivity distribution in orange: across matched-model conditions, a span estimated from one half of the seeds has a mean capture of the other half's error up to $18\%$ versus $0.09\%$ for a random subspace of equal rank. Thus, the observed alignment is not solely an artifact of using the same finite seed set to construct both quantities. Second, we cyclically rotate the field deviations relative to the signed field error in longitude, preserving latitudinal/radial structure while disrupting their longitudinal alignment. Shown in purple is the mean $\mu$ of the error capture computed after the rotations, where the unrotated capture exceeds the mean and lies at the top of the tested distribution in $11/12$ field-condition combinations. The reduced capture indicates that the original longitudinal alignment contributes to the observed overlap beyond generic spatial structure alone. We finally note that the captured energy from the seed-deviation span is a minority of the total error energy but substantially greater than the rank-matched reference expectations (Tables \ref{tab:uq_ecap_ne} and \ref{tab:uq_ecap_temp}).

To summarize, CoroNeRF reports the ensemble mean and cross-seed instability without access to the true field; correlation, partial correlation, sparsification, and error-capture diagnostics require ground truth and are used only to validate this inference-time proxy. In addition, errors shared by ensemble members are absent from cross-seed instability, so consensus is not a correctness guarantee: we characterized this explicitly under a controlled Si abundance mismatch, where localization degrades under the tested range with a pronounced asymmetry between under and over-scaling. Notably, this relative abundance uncertainty can be reduced through co-elemental line selection, although absolute-intensity inversions remain sensitive to the common elemental scale. Finally, we note that the present line set was selected for informative coronal density and temperature response, and conditioning under arbitrary line selections remains future work.
\section{Conclusion} \label{sec:conclusion}
%%%%%%%%%%%%%%%%%%%%%%%%%%%%%%%%%%%%

CoroNeRF jointly reconstructs 3D coronal electron density and temperature through a differentiable multiline atomic-emission renderer. In a controlled synthetic testbed, we show that image fidelity is not a sufficient certificate of thermodynamic-field fidelity and evaluate cross-seed instability as a GT-free-at-inference error-ranking proxy. Additionally, increasing angular coverage substantially mitigates sparse-view geometric degeneracies, while increasing spectral coverage mitigates plasma-thermodynamic degeneracies. Future work will assess generalization and calibration across coronal states and noise realizations, extend the framework to time-dependent and spectropolarimetric observations, and evaluate it on DKIST \citep{Rimmele_2020, Schad_2024} or UCoMP \citep{Landi_2016, Tomczyk_2021} data.

\bibliography{iclr2027_conference}
\bibliographystyle{iclr2027_conference}

\paragraph{AI use statement.} The authors used large language models (ChatGPT and Claude) for finding related works, software development and debugging, checking mathematical rigor, feedback on experimental design and analysis of results, and for editing and proofreading the manuscript. However, all methods, experiments, results, and claims were designed, executed, and verified by the authors, who take full responsibility for the content of this paper.

\paragraph{Reproducibility statement.} The authors conducted this study with reproducibility in mind. Various parts of the forward model (Appendix \ref{app:forward_model}), hyperparameters (Table \ref{tab:hyperparameters}), and benchmark/ablation specifications are detailed in the Appendix of this manuscript. All parameters associated with each experiment, as well as figures in the text, are documented explicitly in configuration code files. Additionally, the full codebase and experiment configuration files are publicly available on the GitHub repository \href{https://github.com/AlanHsu314/CoroNeRF_public}{https://github.com/AlanHsu314/CoroNeRF\_public}.

\clearpage

\appendix

\section{Forward Model and Rendering Details}
\label{app:forward_model}

\subsection{Ground Truth Corona Cube and Emissivity Table Construction}
\label{app:emissivity_table}

Our ground truth coronal model is a MAS (Magnetohydrodynamic Algorithm outside a Sphere) cube from Predictive Science Inc. \citep{Mikic_2007, Lionello_2008}, based on Carrington Rotation 2283 that occurred around April-May 2024. The corona cube is in (longitude $[0, 2\pi]$, colatitude $[0, \pi]$, radius $[1,30]$ R$_\odot$) coordinates with resolution $(299, 142, 154)$. Density and temperature have ranges $\log_{10}(n_{\rm e}/\text{cm}^{-3}) \in [2.16, 12.32]$ and $\log_{10}(T_{\rm e}/\text{K}) \in [5.75, 6.35]$, respectively.

We use the CHIANTI 10.1 atomic database \citep{Dere_1997, Dere_2023} and a Python package \texttt{pyCELP} \citep{Schad_2020} to construct the emissivity tables for our forward model. Our spectral lines are two density-sensitive pairs of forbidden coronal infrared lines: Fe XIII $1075/1080$ nm and Si IX $2585/3935$ nm. While these are the nominal values used in scientific literature, the actual values used in the database are Fe XIII $1074.7/1079.8$ nm and Si IX $2584.6/3929.3$ nm, respectively. When synthesizing these lines, we assume the elemental abundance described in the 2021 Asplund paper \citep{Asplund_2021}.

The emissivity LUT is a function of density $n_{\rm e}$, temperature $T_{\rm e}$, and heliocentric radius $r$, and its outputs are in units of erg s$^{-1}$ cm$^{-3}$ sr$^{-1}$. We retain only the scalar Stokes-I emissivity, neglecting the quadrupole atomic-alignment/depolarization terms associated with the magnetic-field geometry. The LUT axes are chosen from the PSI cube statistics and atomic-response ranges. Queries outside the tabulated $(\log_{10} n_{\rm e}, \log_{10} T_{\rm e}, \log_{10} r)$ domain are clamped to the nearest LUT boundary before interpolation. In particular, the radial LUT spans approximately $1.001$-$10$ R$_\odot$, and LOS samples farther than $10$ R$_\odot$ use the $10$ R$_\odot$ radial boundary response. Our grid resolution is $128 \times 64 \times 32$ in $(\log_{10} n_{\rm e}, \log_{10} T_{\rm e}, \log_{10} r)$. We implement the table in log-space, and we use \texttt{PyTorch} to make it tensor-compatible, notably a tensor-product trilinear interpolator in LUT-coordinate space that uses \texttt{grid\_sample} from \texttt{torch.nn.functional}. The trilinear interpolator additionally has a backstop that returns the nearest boundary even if it is given a value outside the bounds, rather than zeros or a linear extrapolation.

This emissivity LUT is used both in (i) synthetic observation generation using the GT density and temperature values from the PSI cube, and (ii) CoroNeRF for rendering the outputs from the neural plasma field. For (i), we trilinearly interpolate the PSI density and temperature values on a spherical grid using the same \texttt{grid\_sample}, this time with the longitude dimension augmented to support periodic longitude wrapping. 

\subsection{Sampling Geometry and Numerical Quadrature}
\label{app:sampling_quadrature}

CoroNeRF contains objects that live in different coordinate systems. We place the original $[-30, 30]^3$ R$_\odot$ PSI cube centered in a normalized axis-aligned bounding box (AABB) that has bounds $[-1,1]^3$ by normalizing all coordinates by $30$ R$_\odot$. Our ray sampler works in this coordinate system. 

A view is constructed as follows. Given a selected longitude angle, we place a camera at a heliocentric distance of $\approx 215$ R$_\odot$ (Earth-Sun distance), $0^\circ$ latitude, pointed towards the origin. We use a perspective camera model: our rays start from the observer and span a $[-3, 3]$ R$_\odot$ FOV, and each ray forms one pixel on the image plane. At an observer distance of $\approx 215$ R$_\odot$, perspective rays are nearly parallel across the $6$R$_\odot$ FOV. We then mask sightlines with impact parameter $< 1 R_\odot$ to mimic a coronagraph, leaving only a set of coronal pixels to render. Given a pixel to render, we shoot a ray through the AABB, compute the near and far bounds as intersections with the AABB, and compute emissivities at uniform samples along the ray with step size $1/256$ (the number of samples depends on the ray's intersection length with the AABB). To do so at a given sample (in AABB coordinates), we query the neural plasma field $f_{\theta}$ to get density and temperature predictions ($\log_{10} n_{\rm e}, \log_{10} T_{\rm e}$) and use our LUT to compute the emissivity at that point. Note that the LUT takes in the heliocentric radius $r = ||\mathbf{x}||$, computed from physical Cartesian coordinates $\mathbf{x}$. Next, we compute the differential length segment corresponding to each sample point, by scaling our coordinates from AABB to $R_{\odot}$ to cm. We additionally convert the raw LUT emissivity from erg to photon counts, and from steradian$^{-1}$ to arcsec$^{-2}$, so that the integrand $\epsilon_c$ has units photons s$^{-1}$ cm$^{-3}$ arcsec$^{-2}$, and our final intensity observations after integrating along the ray are in units of photons s$^{-1}$ cm$^{-2}$ arcsec$^{-2}$. The LOS integration uses a midpoint Riemann sum. 

\begin{table}[!ht]
	\centering
    \small
    \setlength{\tabcolsep}{3.5pt}
	\begin{tabular}{l|lll} 
		\toprule
		\textbf{step size} & $\Delta s$ (R$_\odot$) & rel-L2 vs finest & $\operatorname{asinhErr}$ vs finest \\
        \midrule
        $1/128$ & $0.234$ & $20.6\%$ & $1.0 \times 10^{-2}$ \\
        $\mathbf{1/256}$ (production) & $0.117$ & $7.5\%$ & $1.7 \times 10^{-3}$ \\
        $1/512$ & $0.059$ & $2.3\%$ & $5.7 \times 10^{-4}$ \\
        $1/1024$ & $0.029$ & | (ref) & | (ref) \\
		\bottomrule
	\end{tabular}
	\caption{Convergence table for selected renderer step sizes. For each step size, we render eight views using the GT cube and compute relative L2 and $\operatorname{asinhErr}$ compared to the finest resolution.} 
	\label{tab:render_stepsize}
	\vspace{-9pt}
\end{table}

We assess the renderer convergence by re-rendering the GT forward model at integration steps $\Delta s \in \{1/128, 1/256, 1/512, 1/1024\}$ (AABB coordinates) on eight views and comparing to the finest resolution (Table \ref{tab:render_stepsize}). The images converge at roughly first-to-second order, and at production step ($1/256 \approx 0.117$ R$_\odot$) they differ from a $4\times$-finer render by $7.5\%$ in relative L2 but only $1.7 \times 10^{-3}$ in the asinh image metric, so production-resolution renders are close to the finer reference under the optimization metric. However, we do not claim discretization-independent field recovery.
Finally, for synthetic dataset generation, the ray sampling and integral procedures are identical, except we use the oracle PSI density and temperature fields instead of the neural plasma field. These fields live on a spherical coordinate system, so we transform AABB to spherical coordinates before sampling for density and temperature values using spherical trilinear field interpolators.

\subsection{LOS visual projections}
\label{app:los_viz}

For visualizing the field, we use rendered field-space projections for intuition, and reserve latitude-longitude plots at fixed shell radii for metric evaluation. 

First, our LOS integration support for visualization is $[1.1, 2.5]$ R$_\odot$. We reuse the ray geometry notation $q$ and define the column density and emissivity-weighted temperature with LOS integrals analogous to Equation \ref{eq:los_eps_integral}:

\begin{align}
    N_{\mathrm{e}}(q) &= \int_{\mathcal{S}(q)} n_{\mathrm{e}} \big( \mathbf{x}_q(t) \big) \, dt , \label{eq:column_density}\\
    \langle T_{\mathrm{e}} \rangle_{\epsilon}(q) &= \frac{\displaystyle \int_{\mathcal{S}(q)} T_{\mathrm{e}}\big( \mathbf{x}_q(t)\big)\, \mathcal{E}\big( \mathbf{x}_q(t) \big) \, dt}{\displaystyle \int_{\mathcal{S}(q)} \mathcal{E}\big( \mathbf{x}_q(t) \big) \, dt} . \label{eq:eps_weight_temperature}
\end{align}

Here, we define a total summed emissivity $\mathcal{E}(\mathbf{x})$ that accounts for all the observed spectral channels

\begin{equation}
    \mathcal{E}(\mathbf{x}) = \sum_{c=1}^C \epsilon_c \big( n_{\mathrm e}(\mathbf{x}), \, T_{\mathrm e}(\mathbf{x}), \, r(\mathbf{x}) \big).
    \label{eq:app_total_eps}
\end{equation}

We choose the column density because it is a real physical quantity: the total electrons accumulated per cm$^{2}$. On the other hand, temperature is not additive, and thus we choose to compute a weighted representative temperature based on the total emission along the LOS.

\subsection{Observational Noise and Forward Model Mismatch}
\label{app:obs_model}

We additionally add heteroscedastic Gaussian noise to our images: for a given channel $c$ and pixel $i$, the observed intensity is drawn from 
\begin{equation}
    I_{i,c}^{\rm obs} \sim \mathcal{N}(I_{i,c}^{\rm true}, \eta\alpha_c I_{i,c}^{\rm true} + \beta_c^2),
    \label{eq:noise_model}
\end{equation}
where $\alpha_c$ is a channel-dependent coefficient for the signal-dependent shot-like variance term, and $\beta_c$ is the background/read-noise floor. We scale the noise by multiplying $\alpha_c$ by $\eta$, a variance multiplier ($\eta = 9$ means a $\times3$ increase in standard deviation). Next, to introduce forward model mismatch, we define the truth $I_{i,c}^{\rm true}$ to be a scaled version of the raw forward model output $I_{i,c}^{\rm raw}$:
\begin{equation}
    I_{i, c}^{\rm true} = \begin{cases}
      a I_{i,c}^{\rm raw} , & c \in \{\text{Si IX channels} \},\\
      I_{i,c}^{\rm raw} , & c \in \{\text{Fe XIII channels} \}.\\
    \end{cases}
\end{equation}

where $a$ is an abundance-mismatch scale applied to the Si IX channels before noise is added. This mismatch models a controlled family of multiplicative channel-wise abundance discrepancies, in particular a relative Fe-to-Si abundance scaling, where $a = 1$ means no forward model error (matched-model). Our implementation scales both Si IX channels by the same value while the Fe XIII channels remain at one, so we refer to $a$ specifically as a Si scale.

In practice, we select $\alpha_c$ such that radial bands across spectral channels achieve approximately the same SNR, reflecting slightly different per-channel noise levels as a real instrument would have. We set the background standard deviation floor to $\beta_c = 0$. For $\eta = 0$ specifically, we bypass this noise generation process entirely. Finally, we note that observation noise is applied to training views only, and held-out image-space evaluation uses corresponding noise-free synthetic views.

%with a numerical clip at $10^{-30}$. In our controlled study, this numerical floor is never active, as the faintest pixel in the corona has $I_{i,c}^{\rm raw} \approx 2.2 \times 10^{-4}$. Our canonical $\alpha_c$ values range from $5 \times 10^{-4}$ to $5 \times 10^{-2}$, so $\eta\alpha_c I_{i,c}^{\rm true}$ values lie above the numerical floor. For $\eta = 0$ specifically, we bypass this noise generation process so the numerical floor is not applied.

%This is especially important for our linear identifiability analysis, as an incorrect floor will make park pixels deceptively informative (Table \ref{tab:uq_floor_sweep} shows that our identifiability diagnostics are insensitive to different noise floors).

\subsection{Multiresolution Hash Grid Model}
\label{app:hash_grid}
% go through multired hash grids, and ablations

Our multiresolution hash-grid representation is based on the encoding described by \citep{Muller_2022}, where we encode spatial information as trainable grid features, and then separate decoder MLP heads combine the features into density and temperature estimates. The grids themselves do not store features directly at each spatial point, as this does not scale with resolution. The grid is indexed by hashing the spatial coordinate of the point, where we define the size as $2^G$ by $F$: $G$ is the hashmap table size in $\log_2$, and $F$ is the number of features. The full hash grid contains many such levels, each associated with a physical spatial resolution. Given a sample position, the features at that point are computed by trilinearly interpolating the features at the eight corners of the cube surrounding that point. The positions of those eight corners are computed relative to the fixed spatial resolution of that level. The coordinates are then hashed and used in the trainable hashmap to retrieve the features.

For a multiresolution hash grid with $L$ levels, we define the coarsest (base) resolution as $N_0$ and the finest resolution as $N_{L-1}$. We then spread the resolution geometrically, where the resolution at level $\ell$ is given by

\begin{equation}
    N_\ell = N_0 b^\ell \, , \quad b = \exp \left( \frac{\ln (N_{L-1}/N_0)}{L-1} \right).
\end{equation}

Our hash function is a bitwise $\operatorname{XOR}$ hash. While hashing integer coordinates may introduce many collisions, we use multiple levels, each with a different physical spatial resolution. Thus, the multiresolution nature of our grid not only encodes information at different spatial levels, but also mitigates the effect of collisions once we concatenate the features at all levels. 

In our implementation, each level has a hash table with $2^{20}$ entries and $F=2$ features per entry, with a total of $L=20$ levels that span a spatial resolution going from $N_0 = 16$ (coarsest) to $N_{L-1} = 512$ (finest). Our decoder heads are each two hidden layers of width $64$.

During training (per minibatch of rays), because our forward model is differentiable with respect to the neural plasma field (and thus the model parameters), the gradients first backpropagate through the differentiable line-emission renderer, and then through the emissivity lookup table to the predicted $\log_{10} n_{\rm{e}}$ and $\log_{10}T_{\rm{e}}$ fields, and finally to the multiresolution hash grid. The lookup table itself and ray geometry and sampling are fixed during optimization.

\clearpage

\section{Metrics}
\label{app:metrics}

Our evaluations are conducted in two spaces. \emph{Image-space metrics} compare rendered line-intensity images to held-out observations. \emph{Field-space metrics} compare the reconstructed density and temperature fields to their ground-truth counterparts on spherical shells. Field-space metrics are the primary metrics in this work as the goal is a latent joint physical-field recovery rather than image synthesis.

\paragraph{Notation.}
For any finite index set $\mathcal{A}$, we write 
\begin{equation}
    \langle f(a) \rangle_{a \in \mathcal{A}} = \frac{1}{|\mathcal{A}|} \sum_{a\in\mathcal{A}} f(a)
\end{equation}

for the empirical average. We also use the same indexing as in Section \ref{sec:methods} of the main text: $v \in \mathcal{V} = \{1, \ldots, V\}$ indexes viewpoints, $p\in\Omega_v$ indexes valid pixels in view $v$, and $c \in \mathcal{C} = \{1 , \ldots, C \}$ indexes spectral line channels.

\subsection{Image-Space Metrics}

Let $I_{v, p, c}$ and $\hat{I}_{v, p, c}$ denote the observed and predicted line intensities for view $v$, pixel $p$, and channel $c$. The per-view, per-channel mean-squared error is 

\begin{equation}
    \text{MSE}_{v,c}(I) = \left\langle (\hat{I}_{v, p, c} - I_{v, p, c})^2 \right\rangle_{p\in\Omega_v}.
\end{equation}

The per-view, per-channel PSNR is

\begin{equation}
    \text{PSNR}_{v,c}(I) = 10 \log_{10} \left( \frac{R_{v,c}^2}{\text{MSE}_{v,c}(I)} \right),
\end{equation}

where $R_{v,c}$ is the $1$st-to-$99$th-percentile target-intensity range for that view and channel, with max-minus-min as a fallback if the percentile range is degenerate. Dataset-level image-space metrics are obtained by averaging over views and channels:

\begin{equation}
    \text{PSNR}(I) = \left\langle \text{PSNR}_{v,c}(I) \right\rangle_{v\in \mathcal{V},c\in\mathcal{C}}.
\end{equation}

Next, our fixed-scale asinh image error is defined as follows:

\begin{equation}
    \text{asinhErr}(I)=\left\langle\left| \operatorname{asinh}\!\left(\frac{\hat I_{v,p,c}}{s_c}\right) -\operatorname{asinh}\!\left(\frac{I_{v,p,c}}{s_c}\right)\right|\right\rangle_{p \in \Omega_v,\; v \in \mathcal{V}, \; c \in \mathcal{C}},
\end{equation}

where $s_c$ is the channel-dependent intensity scale specified by the evaluation protocol (in the noise-view study, held-out $\text{asinhErr}$ uses the common base per-channel scales). This metric is well-defined for signed intensities and behaves logarithmically for large intensity magnitudes.

\subsection{Field-Space Metrics}
\label{app:field_metrics}

Field metrics are computed on spherical shells. Let $r\in\mathcal{R}$ index evaluation shell radii, and let $s\in\mathcal{S}_r$ index valid samples on the shell at radius $r$. We use $m$ to denote a density or temperature log-field
\begin{equation}
    m \in \{ \log_{10}n_{\rm e}, \; \log_{10} T_{\rm e} \}.
\end{equation}

Let $m_{r,s}$ and $\hat{m}_{\theta,r,s}$ be the ground-truth and predicted log-field values. For a radial band $\mathcal{R}_{\text{band}}\subseteq \mathcal{R}$, we define the mean absolute log error

\begin{equation}
    \text{MAE}_{\text{band}}(m) = \left\langle |\hat{m}_{\theta, r, s} - m_{r,s}| \right\rangle_{s \in \mathcal{S}_r, \; r \in \mathcal{R}_{\text{band}}}.
\end{equation}

We also report the absolute relative error

\begin{equation}
    \text{AbsRel}_{\text{band}}(m) = \left\langle \frac{|10^{\hat{m}_{\theta, r, s}} - 10^{m_{r,s}}|}{\operatorname{max}(|10^{m_{r,s}}|,\epsilon_0)} \right\rangle_{s \in \mathcal{S}_r, \; r \in \mathcal{R}_{\text{band}}},
\end{equation}

where $\epsilon_0 = 10^{-30}$ is a small numerical floor.

Finally, for model-mismatch studies, the sign of the error is also important, so we additionally report signed shell mean error:

\begin{equation}
    \text{ME}_{\text{band}}(m) = \left\langle \hat{m}_{\theta, r, s} - m_{r,s} \right\rangle_{s \in \mathcal{S}_r, \; r \in \mathcal{R}_{\text{band}}}.
\end{equation}

Unlike absolute error metrics, this signed metric indicates whether a reconstructed field systematically overestimates or underestimates the ground truth field.

\section{Ensemble Uncertainty Quantification}

\subsection{Cross-Seed Instability and Correlation Analysis}
\label{app:cross_seed}

We denote a plasma quantity as $m \in \{ \log_{10}n_{\rm e}, \log_{10} T_{\rm e}\}$. Every definition below is per-$m$ for readability (fields are in dex), and voxels are indexed by position $\mathbf{x}$. Let us reuse the shell and radial notation from Appendix \ref{app:metrics}: $\mathcal{S}_r = \{\mathbf{x}: r(\mathbf{x}) = r \}$ as a radial shell, and $\mathcal{R}_{[r_1, r_2]} = \{ \mathbf{x}: r_1 \le r(\mathbf{x}) \le r_2 \}$ as a radial band from $r_1$ to $r_2$. Finally, let $ m^*(\mathbf{x})$ be the ground truth discretized field.

We train $K$ models, with fixed model architecture, observational protocol $(\eta, V, \Lambda, a)$, and training dataset, only varying the seeds $k = 1\dots K$ \citep{Lakshminarayanan_2017}. In particular, seed $k$ gives a predicted field $\hat{m}_k(\mathbf{x})$. Define the ensemble mean and cross-seed instability across the seeds, respectively, as follows:

\begin{align}
	\overline{m}(\mathbf{x}) &= \frac{1}{K} \sum_k \hat{m}_k(\mathbf{x}), \label{eq:app_ensemble_mean} \\
	\sigma_{\rm ens}(\mathbf{x}) &= \left[ \frac{1}{K-1}\sum_k (\hat{m}_k(\mathbf{x}) - \overline{m}(\mathbf{x}))^2 \right]^{1/2}.
	\label{eq:app_ensemble_disagreement}
\end{align}

$\sigma_{\rm ens}$ is a GT-free-at-inference empirical metric that is a field-error localization proxy. Large $\sigma_{\rm ens}$ identifies regions where the recovered field is sensitive to training seeds under the fixed pipeline, while small $\sigma_{\rm ens}$ identifies regions of cross-seed consensus. On the other hand, the absolute ensemble mean error, which can never be known without access to the GT field $m^*$, quantifies the error of the ensemble-mean field:

\begin{equation}
    \epsilon_{\rm mean}(\mathbf{x}) = |\overline{m}(\mathbf{x}) - m^*(\mathbf{x})|. 
    \label{eq:app_ensemble_mean_error}
\end{equation}

Note that $\epsilon_{\rm mean}$ is not a statistical bias, which would require an expectation over multiple noise realizations. 

\begin{figure}[!th]
	\begin{center}
		\includegraphics[width=1\linewidth]{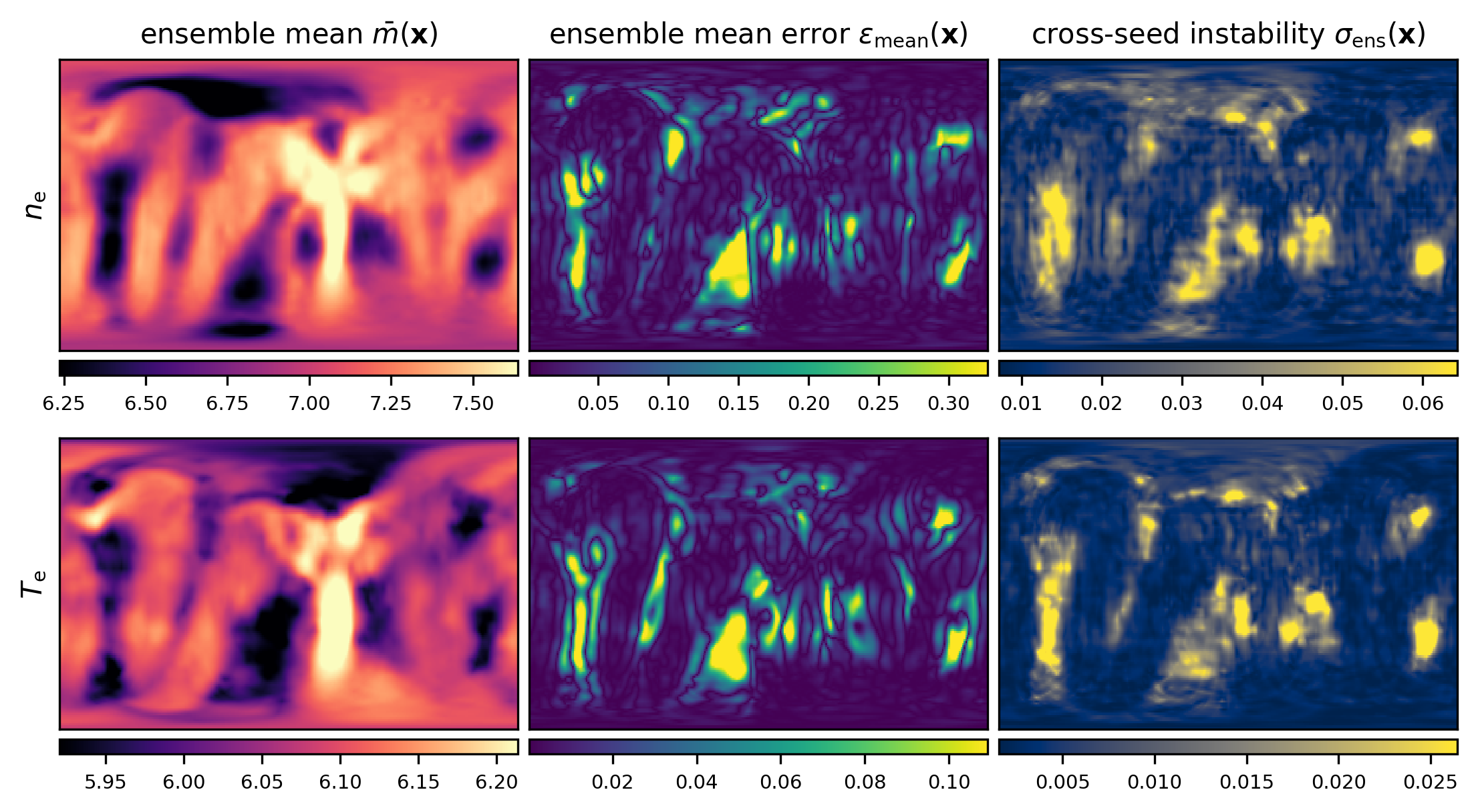}
	\end{center}
	\caption{Ensemble reconstruction panels of the 3D density (top) and temperature (bottom) fields at $1.51$R$_\odot$ in the high-noise regime ($\eta = 9, V = 300, \text{all four lines}, a = 1$). We show the ensemble mean $\overline{m}(\mathbf{x})$ (Equation \ref{eq:app_ensemble_mean}), ensemble-mean error $\epsilon_{\rm mean}(\mathbf{x})$ (unobservable, Equation \ref{eq:app_ensemble_mean_error}), and cross-seed instability $\sigma_{\rm ens}(\mathbf{x})$ (Equation \ref{eq:app_ensemble_disagreement}).}
	\label{fig:benchF_seed_disagreement_app}
\end{figure}

Finally, to quantitatively express localization correlation, let us use error target $\epsilon_{\rm mean}$ and spatial domain $\Omega \in \{\mathcal{S}_r, \mathcal{R}_{[r_1, r_2]} \}$. Define the Spearman rank correlation coefficient between the cross-seed instability and the error over $\Omega$ as
\begin{equation}
	\rho_{\epsilon} = \operatorname{Spearman}_{x\in\Omega}(\sigma_{\rm ens}, \epsilon_{\rm mean}) = \operatorname{corr}(\operatorname{rank} \sigma_{\rm ens}, \operatorname{rank} \epsilon_{\rm mean}).
    \label{eq:app_spearman_corr}
\end{equation}

We note that our ensemble-localization study uses $300$ evenly-spaced training views with a disjoint $30$-view interleaved evenly-spaced holdout. All members within each condition share the same split, and for noisy conditions they share the same fixed noise realization. Figure \ref{fig:benchF_seed_disagreement_app} shows sample ensemble mean $\overline{m}(\mathbf{x})$, ensemble-mean error $\epsilon_{\rm mean}(\mathbf{x})$, and cross-seed instability $\sigma_{\rm ens}(\mathbf{x})$ panels for our canonical noisy condition.

Next, our baseline signal-strength proxy is the inverse-emissivity proxy, given by $b_{\mathcal{E}} = 1/\mathcal{E}$, where $\mathcal{E} > 0$ is the total local emissivity summed over all spectral channels (Equation \ref{eq:app_total_eps}). We use this as a baseline signal-strength confound: low-emissivity regions are a plausible confound because they are more weakly constrained and may exhibit larger reconstruction error. We note that in mismatch conditions we intentionally retain the nominal, unscaled GT emissivity, so $b_\mathcal{E}$ remains an oracle probe of the underlying physical signal rather than an observation-consistent proxy.

We then compute per-condition pairwise correlations between the cross-seed instability $\sigma_{\rm ens}$, the ensemble mean error $\epsilon_{\rm mean}$, and our emissivity proxy $b_\mathcal{E}$, to answer the following question: does the rank association between $\sigma_{\rm ens}$ and $\epsilon_{\rm mean}$ remain after controlling for the oracle inverse-emissivity proxy? To do so, we control for $b_{\mathcal{E}}$ by rank-transforming each variable and computing the standard partial Pearson correlation on those ranks:

\begin{equation}
    \rho(\sigma_{\rm ens}, \epsilon_{\rm mean} \, | \, b_{\mathcal{E}}) = \frac{\rho(\sigma_{\rm ens}, \epsilon_{\rm mean}) - \rho(\sigma_{\rm ens}, b_\mathcal{E})\rho(b_\mathcal{E}, \epsilon_{\rm mean})}{\sqrt{(1 - \rho^2(\sigma_{\rm ens}, b_\mathcal{E}))(1-\rho^2(b_\mathcal{E}, \epsilon_{\rm mean}))}}.
\end{equation}

A strong positive partial correlation indicates a nontrivial linear association between the rank-transformed $\sigma_{\rm ens}$ and $\epsilon_{\rm mean}$, after linear adjustment for the rank-transformed $b_\mathcal{E}$ proxy (in other words, suggesting that $\sigma_{\rm ens}$ captures error structure beyond what is explained by signal strength alone).

\begin{table}[!ht]
	\centering
    %\small
    %\setlength{\tabcolsep}{3.5pt}
    \renewcommand{\arraystretch}{1.5}
	\begin{tabular}{l|llll} 
		\toprule
		\textbf{Condition} & $\rho(\sigma_{\rm ens}, \epsilon_{\rm mean})$ $\uparrow$ & $\rho(b_\mathcal{E}, \epsilon_{\rm mean})$ & partial $\rho(\sigma_{\rm ens}, \epsilon_{\rm mean} \, | \, b_\mathcal{E})$ $\uparrow$ & $\rho(\sigma_{\rm ens}, b_\mathcal{E})$ \\
        \midrule
        noiseless ($300$v) & $0.761_{-0.029}^{+0.022}$ & $0.072_{-0.058}^{+0.052}$ & 	$0.759_{-0.027}^{+0.022}$ & $0.081$ \\
        $100$ views & $0.773_{-0.027}^{+0.021}$ & $0.085_{-0.056}^{+0.050}$ & $0.772_{-0.025}^{+0.020}$ &  $0.086$ \\

        \midrule
        
        $20$ views & $0.778_{-0.034}^{+0.029}$ & $-0.019_{-0.056}^{+0.045}$ & $0.780_{-0.035}^{+0.029}$ & $0.038$ \\
        $5$ views & $0.712_{-0.050}^{+0.042}$ & $-0.103_{-0.094}^{+0.088}$ & 	$0.708_{-0.051}^{+0.043}$ & $-0.161$ \\

        \midrule
        
        Noise $\times9$ & $0.572_{-0.052}^{+0.041}$ & $0.405_{-0.042}^{+0.038}$ & $0.447_{-0.059}^{+0.052}$ & $0.621$ \\
        Noise $\times25$ & $0.555_{-0.055}^{+0.039}$ & $0.438_{-0.044}^{+0.036}$ & $0.393_{-0.059}^{+0.052}$ & $0.663$ \\

        \midrule

        Si $\times0.6$ & $0.438_{-0.082}^{+0.073}$ & $0.167_{-0.063}^{+0.060}$ & $0.417_{-0.087}^{+0.079}$ & $0.229$ \\
        Si $\times0.8$ & $0.387_{-0.071}^{+0.067}$ & $0.209_{-0.068}^{+0.056}$ & $0.359_{-0.070}^{+0.069}$ & $0.207$ \\
        Si $\times1.2$ & $0.166_{-0.071}^{+0.067}$ & $0.435_{-0.061}^{+0.055}$ & $0.070_{-0.061}^{+0.061}$ & $0.243$ \\
        Si $\times1.4$ & $-0.017_{-0.087}^{+0.081}$ & $0.476_{-0.055}^{+0.049}$ & $-0.082_{-0.079}^{+0.070}$ & $0.115$ \\
		\bottomrule
	\end{tabular}
	\caption{(Density) Pairwise Spearman correlations between cross-seed instability $\sigma_{\rm ens}$, ensemble mean error $\epsilon_{\rm mean}$, and emissivity proxy $b_\mathcal{E}$.} 
	\label{tab:uq_correlations_ne}
	%\vspace{-9pt}
\end{table}

\begin{table}[!ht]
	\centering
    %\small
    %\setlength{\tabcolsep}{3.5pt}
    \renewcommand{\arraystretch}{1.5}
	\begin{tabular}{l|llll} 
		\toprule
		\textbf{Condition} & $\rho(\sigma_{\rm ens}, \epsilon_{\rm mean})$ $\uparrow$ & $\rho(b_\mathcal{E}, \epsilon_{\rm mean})$ & partial $\rho(\sigma_{\rm ens}, \epsilon_{\rm mean} \, | \, b_\mathcal{E})$ $\uparrow$ & $\rho(\sigma_{\rm ens}, b_\mathcal{E})$ \\
        \midrule
        noiseless ($300$v) & $0.726_{-0.027}^{+0.025}$ & $0.001_{-0.066}^{+0.067}$ & $0.732_{-0.027}^{+0.024}$ & $0.122$ \\
        $100$ views & $0.732_{-0.023}^{+0.019}$ & $0.014_{-0.068}^{+0.063}$ & $0.735_{-0.022}^{+0.019}$ &  $0.115$ \\

        \midrule
        
        $20$ views & $0.717_{-0.034}^{+0.031}$ & $0.092_{-0.055}^{+0.053}$ & $0.716_{-0.032}^{+0.029}$ & $0.061$ \\
        $5$ views & $0.611_{-0.041}^{+0.034}$ & $0.015_{-0.057}^{+0.049}$ & 	$0.626_{-0.040}^{+0.032}$ & $-0.189$ \\

        \midrule
        
        Noise $\times9$ & $0.590_{-0.046}^{+0.044}$ & $0.325_{-0.052}^{+0.046}$ & $0.521_{-0.049}^{+0.046}$ & $0.579$ \\
        Noise $\times25$ & $0.601_{-0.049}^{+0.042}$ & $0.382_{-0.051}^{+0.045}$ & $0.503_{-0.058}^{+0.047}$ & $0.624$ \\

        \midrule

        Si $\times0.6$ & $0.424_{-0.073}^{+0.068}$ & $0.317_{-0.067}^{+0.058}$ & $0.361_{-0.072}^{+0.075}$ & $0.312$ \\
        Si $\times0.8$ & $0.448_{-0.065}^{+0.062}$ & $0.368_{-0.052}^{+0.046}$ & $0.365_{-0.064}^{+0.065}$ & $0.356$ \\
        Si $\times1.2$ & $0.250_{-0.071}^{+0.068}$ & $0.228_{-0.069}^{+0.069}$ & $0.226_{-0.065}^{+0.067}$ & $0.141$ \\
        Si $\times1.4$ & $0.096_{-0.063}^{+0.069}$ & $0.138_{-0.066}^{+0.066}$ & $0.101_{-0.057}^{+0.063}$ & $-0.027$ \\
		\bottomrule
	\end{tabular}
	\caption{(Temperature) Pairwise Spearman correlations between cross-seed instability $\sigma_{\rm ens}$, ensemble mean error $\epsilon_{\rm mean}$, and emissivity proxy $b_\mathcal{E}$.} 
	\label{tab:uq_correlations_T}
	%\vspace{-9pt}
\end{table}

Tables \ref{tab:uq_correlations_ne} (density) and \ref{tab:uq_correlations_T} (temperature) show pairwise correlations between $\sigma_{\rm ens}$, $\epsilon_{\rm mean}$, and $b_\mathcal{E}$, as well as the partial correlation $\rho(\sigma_{\rm ens}, \epsilon_{\rm mean} \, | \, b_\mathcal{E})$, in four different regimes: noiseless (300v, 100v), sparse-view (20v, 5v), high-noise ($\times9$, $\times25$), and forward model abundance mismatch (Si $\times0.6$, $\times0.8$, $\times1.2$, $\times1.4$). 

We evaluate the diagnostics on a $30\times18\times8$ longitude-latitude-radius grid, with the radial coordinates spanning $1.05$-$3$ R$_\odot$. To assess within-scene resampling variability, we treat each longitude slice, containing its associated latitude-radius samples, as one cluster. Resampling entire slices preserves their internal spatial dependence but does not explicitly model dependence between neighboring longitude slices. For each of $2000$ bootstrap replicates, we draw $30$ longitude indices independently with replacement from the $30$ available indices and concatenate the corresponding slices. The same sampled indices are applied to $\sigma_{\rm ens}$, $\epsilon_{\rm mean}$, and $b_\mathcal{E}$. We recompute the correlation, partial correlation, and $\operatorname{nAUSE}$ (Appendix \ref{app:sparsification}) on each resample and report the $2.5$th and $97.5$th percentiles. Finally, correlation and sparsification calculations retain only samples in the inner radial band, and all statistics use unweighted samples rather than spherical area or volume weighting.
\begin{table}[!ht]
	\centering
    %\small
    %\setlength{\tabcolsep}{3.5pt}
    \renewcommand{\arraystretch}{1.5}
	\begin{tabular}{l|llll} 
		\toprule
		\textbf{Condition} & pooled & $r=1.33$ R$_\odot$ & $r=1.61$ R$_\odot$ & $r=1.89$ R$_\odot$ \\
        \midrule
        noiseless ($300$v) & $0.76$ [$0.76$] & $0.74$ [$0.74$] & $0.77$ [$0.75$] & $0.76$ [$0.75$]\\
        $100$ views & $0.77$ [$0.77$] & $0.75$ [$0.75$] & $0.78$ [$0.77$] & $0.77$ [$0.75$] \\

        \midrule
        
        $20$ views & $0.78$ [$0.78$] & $0.78$ [$0.78$] & $0.78$ [$0.78$] & $0.76$ [$0.73$] \\
        $5$ views & $0.71$ [$0.71$] & $0.68$ [$0.68$] & $0.75$ [$0.75$] & $0.61$ [$0.60$] \\

        \midrule
        
        Noise $\times9$ & $0.57$ [$0.45$] & $0.59$ [$0.58$] & $0.54$ [$0.44$] & $0.41$ [$0.20$] \\
        Noise $\times25$ & $0.56$ [$0.39$] & $0.53$ [$0.51$] & $0.54$ [$0.41$] & $0.38$ [$0.15$] \\

        \midrule

        Si $\times0.6$ & $0.44$ [$0.42$] & $0.47$ [$0.50$] & $0.46$ [$0.44$] & $0.37$ [$0.33$] \\
        Si $\times0.8$ & $0.39$ [$0.36$] & $0.39$ [$0.43$] & $0.40$ [$0.37$] & $0.37$ [$0.28$] \\
        Si $\times1.2$ & $0.17$ [$0.07$] & $0.12$ [$0.14$] & $0.13$ [$0.09$] & $0.17$ [$0.08$] \\
        Si $\times1.4$ & $-0.02$ [$-0.08$] & $-0.09$ [$-0.02$] & $-0.04$ [$-0.01$] & $-0.04$ [$-0.05$] \\
		\bottomrule
	\end{tabular}
	\caption{(Density) Per-shell correlation $\rho(\sigma_{\rm ens}, \epsilon_{\rm mean})$ (partial $\rho(\sigma_{\rm ens}, \epsilon_{\rm mean} \, | \, b_\mathcal{E})$ in brackets), for selected conditions.} 
	\label{tab:uq_pershell_correlations_ne}
	%\vspace{-9pt}
\end{table}

\begin{table}[!ht]
	\centering
    %\small
    %\setlength{\tabcolsep}{3.5pt}
    \renewcommand{\arraystretch}{1.5}
	\begin{tabular}{l|llll} 
		\toprule
		\textbf{Condition} & pooled & $r=1.33$ R$_\odot$ & $r=1.61$ R$_\odot$ & $r=1.89$ R$_\odot$ \\
        \midrule
        noiseless ($300$v) & $0.73$ [$0.73$] & $0.70$ [$0.70$] & $0.72$ [$0.70$] & $0.77$ [$0.75$]\\
        $100$ views & $0.73$ [$0.74$] & $0.73$ [$0.73$] & $0.70$ [$0.68$] & $0.76$ [$0.75$] \\

        \midrule
        
        $20$ views & $0.72$ [$0.72$] & $0.72$ [$0.73$] & $0.67$ [$0.65$] & $0.75$ [$0.72$] \\
        $5$ views & $0.61$ [$0.63$] & $0.70$ [$0.72$] & $0.50$ [$0.50$] & $0.53$ [$0.51$] \\

        \midrule
        
        Noise $\times9$ & $0.59$ [$0.52$] & $0.66$ [$0.66$] & $0.54$ [$0.45$] & $0.49$ [$0.30$] \\
        Noise $\times25$ & $0.60$ [$0.50$] & $0.61$ [$0.61$] & $0.55$ [$0.44$] & $0.50$ [$0.29$] \\

        \midrule

        Si $\times0.6$ & $0.42$ [$0.36$] & $0.30$ [$0.31$] & $0.38$ [$0.30$] & $0.50$ [$0.43$] \\
        Si $\times0.8$ & $0.45$ [$0.36$] & $0.35$ [$0.35$] & $0.43$ [$0.30$] & $0.52$ [$0.38$] \\
        Si $\times1.2$ & $0.25$ [$0.23$] & $0.30$ [$0.33$] & $0.18$ [$0.19$] & $0.24$ [$0.21$] \\
        Si $\times1.4$ & $0.10$ [$0.10$] & $0.14$ [$0.20$] & $0.08$ [$0.12$] & $0.06$ [$0.07$] \\
		\bottomrule
	\end{tabular}
	\caption{(Temperature) Per-shell correlation $\rho(\sigma_{\rm ens}, \epsilon_{\rm mean})$ (partial $\rho(\sigma_{\rm ens}, \epsilon_{\rm mean} \, | \, b_\mathcal{E})$ in brackets), for selected conditions.} 
	\label{tab:uq_pershell_correlations_temp}
	%\vspace{-9pt}
\end{table}

In Tables \ref{tab:uq_pershell_correlations_ne} (density) and \ref{tab:uq_pershell_correlations_temp} (temperature), we show within-shell Spearman correlation $\rho(\sigma_{\rm ens}, \epsilon_{\rm mean})$ between cross-seed instability and absolute ensemble mean error, with the partial correlation $\rho(\sigma_{\rm ens}, \epsilon_{\rm mean} \, | \, b_\mathcal{E})$ in brackets. We compute these correlations for 3 shells (columns) at R$ = [1.33, 1.61, 1.89]$ R$_\odot$, each with 540 voxels on a controlled grid. Within-shell values track the pooled correlation for the selected conditions, showing that the correlation is not solely explained by a radial trend.

\subsection{Sparsification Analysis}
\label{app:sparsification}

A useful error-ranking proxy should allow the average error to improve by systematically discarding voxels it flags as least trustworthy (sparsification process). A metric we use is the Area Under the Sparsification Error curve ($\operatorname{AUSE}$, \cite{Ilg_2018, Poggi_2020}), a standard protocol used to assess neural-field uncertainty quantification \citep{Shen_2022, Sunderhauf_2023, Goli_2024}. To construct this, we pick a ranking scalar $z$ (such as $\sigma_{\rm ens}$, $b_{\mathcal{E}}$, $\epsilon_{\rm mean}$ (oracle), random), and sort all voxels by decreasing $z$. We note that the random baseline is the exact expected uniform-retention curve (the global mean error), not a finite permutation average. For a total of $N$ voxels, we then parameterize the process of removing voxels via $K_z(t)$, the subset retaining approximately the lowest-ranked $(1-t)$ fraction of voxels. We then define the risk as the mean error over the kept set
\begin{equation}
    \operatorname{Risk}_z(t) = \frac{1}{|K_z(t)|} \sum_{i \in K_z(t)} \epsilon_{\rm{mean},i}.
\end{equation}

Plotting $\operatorname{Risk}_z(t)$ against $t$ gives the sparsification curve. At $t = 0$, all curves retain all voxels. The oracle $z = \epsilon_{\rm mean}$ decreases the fastest as it removes the largest errors first, while a random ranking keeps the MAE relatively flat. A useful ranking proxy should approach the oracle, while arbitrary proxies may perform similar to or worse than the random baseline.

To summarize each curve with a single metric, we first compute the sparsification error
\begin{equation}
    \operatorname{SE}_z(t) = \operatorname{Risk}_z(t) - \operatorname{Risk}_{\rm oracle}(t) \ge 0,
\end{equation}

which is the gap to the oracle at each level, and then compute the area under the curve
\begin{equation}
    \operatorname{AUSE}_z = \int_0^{t_{\rm max}} \operatorname{SE}_z(t) dt,
\end{equation}

where we integrate up to $t_{\rm max} = 0.95$ to avoid a noisy terminal spike when few voxels remain. In practice, this is done using a discretized $t$-grid. A lower $\operatorname{AUSE}$ indicates a ranking closer to that of the oracle ($\operatorname{AUSE} = 0$). Finally, to compare across conditions, we normalize the $\operatorname{AUSE}$ by a random ranking, such that $0$ gives perfect, while $1$ gives no-better-than-random, and $>1$ gives worse-than-random:
\begin{equation}
    \operatorname{nAUSE}_z = \frac{\operatorname{AUSE_z}}{\operatorname{AUSE}_{\rm random}}.
    \label{eq:app_nAUSE}
\end{equation}

\begin{figure}[!th]
	\begin{center}
		\includegraphics[width=1\linewidth]{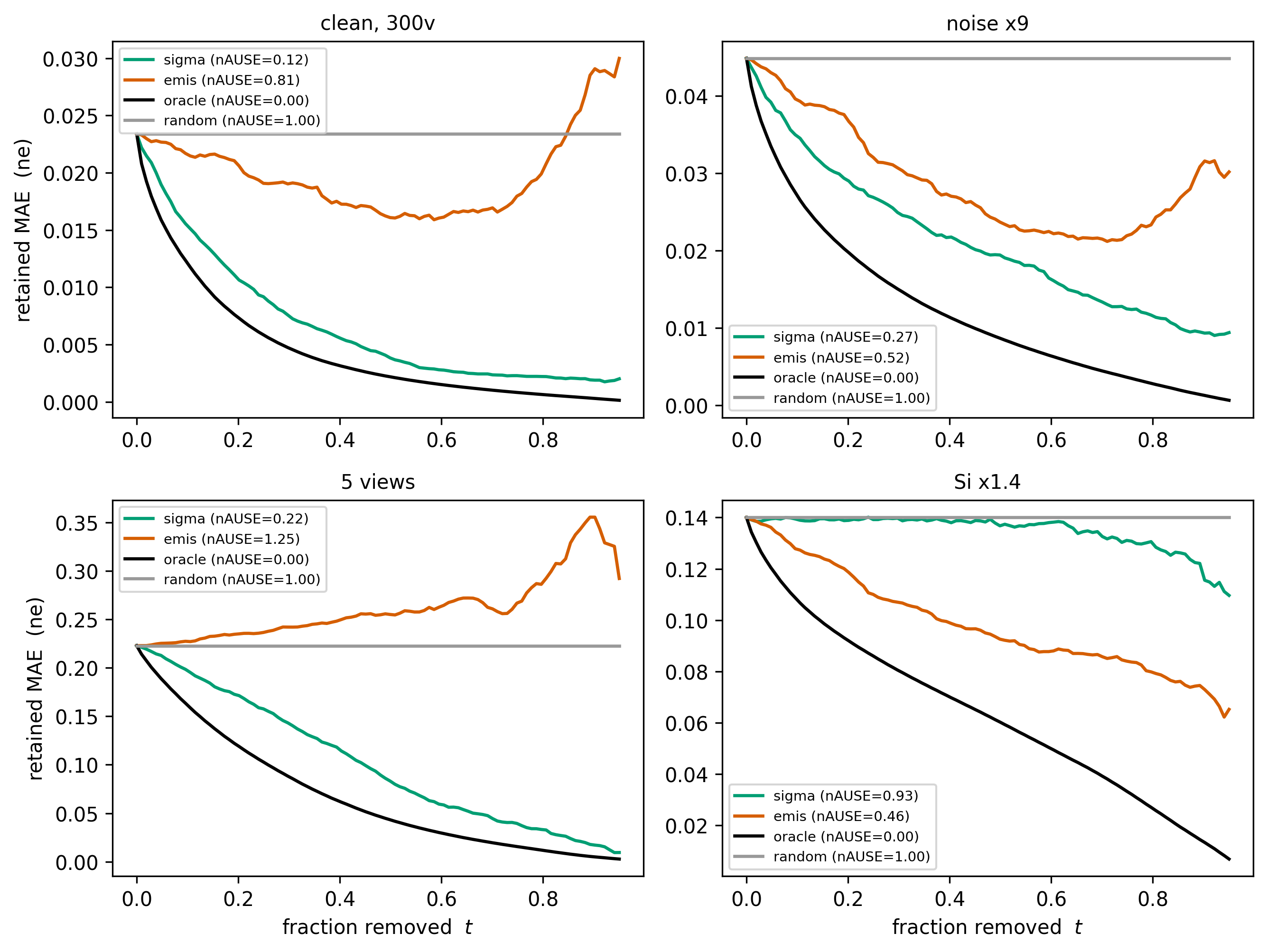}
	\end{center}
	\caption{Density-error sparsification curves for the noiseless, high-noise, sparse-view, and Si-mismatch regimes. Lower remaining MAE is better, and the random reference is the exact expected uniform-retention curve.}
	\label{fig:uq_sparsification_curves}
\end{figure}

Figure \ref{fig:uq_sparsification_curves} shows the sparsification curves for four main regimes: noiseless, high-noise, sparse-view, and forward-model mismatch. In each plot, we show curves for various ranking scalars $z$: the cross-seed instability (green), emissivity proxy (orange), oracle (black), and random (gray). We find that across the matched-model noiseless, high-noise, and sparse-view regimes, the cross-seed instability tracks the oracle closely and much better than our physical emissivity proxy, and degrades towards random ranking under the strongest tested forward-model mismatch. Tables \ref{tab:uq_sparsification_ne} (density) and \ref{tab:uq_sparsification_T} (temperature) tabulate the values in the curves, along with the raw $\operatorname{AUSE}$ values. The $\operatorname{nAUSE}$ intervals are $95\%$ within-scene longitude-slice bootstrap intervals using the same $2000$-resample procedure. Note that raw $\operatorname{AUSE}$ combines the scale of reconstruction error with ranking quality, so we therefore use $\operatorname{nAUSE}$ for comparisons of relative ranking performance across conditions and report raw $\operatorname{AUSE}$ for reference.

\begin{table}[!ht]
	\centering
    \small
    \setlength{\tabcolsep}{3.5pt}
    \renewcommand{\arraystretch}{1.5}
	\begin{tabular}{l|llll} 
		\toprule
		\textbf{Condition} & $\operatorname{AUSE}_\sigma$ & $\operatorname{AUSE}_b$ & $\operatorname{nAUSE}_\sigma$ & $\operatorname{nAUSE}_b$ \\
        \midrule
        noiseless ($300$v) & $0.0021$ & $0.0145$ & $0.115_{-0.012}^{+0.019}$ & $0.809_{-0.096}^{+0.120}$ \\
        $100$ views & $0.0021$ & $0.0147$ & $0.113_{-0.012}^{+0.016}$ & $0.794_{-0.094}^{+0.115}$ \\

        \midrule
        
        $20$ views & $0.0062$ & $0.0657$ & $0.117_{-0.020}^{+0.026}$ & $1.239_{-0.102}^{+0.101}$ \\
        $5$ views & $0.0327$ & $0.1834$ & 	$0.222_{-0.035}^{+0.043}$ & $1.245_{-0.135}^{+0.139}$ \\

        \midrule
        
        Noise $\times9$ & $0.0085$ & $0.0161$ & $0.274_{-0.036}^{+0.047}$ & $0.520_{-0.063}^{+0.075}$ \\
        Noise $\times25$ & $0.0108$ & $0.0168$ & $0.304_{-0.044}^{+0.056}$ & $0.472_{-0.055}^{+0.067}$ \\

        \midrule

        Si $\times0.6$ & $0.0676$ & $0.1383$ & $0.408_{-0.062}^{+0.074}$ & $0.834_{-0.097}^{+0.102}$ \\
        Si $\times0.8$ & $0.0312$ & $0.0579$ & $0.416_{-0.050}^{+0.060}$ & $0.772_{-0.096}^{+0.107}$ \\
        Si $\times1.2$ & $0.0302$ & $0.0227$ & $0.610_{-0.060}^{+0.072}$ & $0.459_{-0.047}^{+0.063}$ \\
        Si $\times1.4$ & $0.0676$ & $0.0335$ & $0.932_{-0.084}^{+0.098}$ & $0.462_{-0.046}^{+0.060}$ \\
		\bottomrule
	\end{tabular}
	\caption{(Density) Sparsification summary for selected ranking metrics across selected conditions.} 
	\label{tab:uq_sparsification_ne}
	%\vspace{-9pt}
\end{table}

\begin{table}[!ht]
	\centering
    \small
    \setlength{\tabcolsep}{3.5pt}
    \renewcommand{\arraystretch}{1.5}
	\begin{tabular}{l|llll} 
		\toprule
		\textbf{Condition} & $\operatorname{AUSE}_\sigma$ & $\operatorname{AUSE}_b$ &  $\operatorname{nAUSE}_\sigma$ & $\operatorname{nAUSE}_b$ \\
        \midrule
        noiseless ($300$v) & $0.0008$ & $0.0052$ & $0.140_{-0.025}^{+0.028}$ & $0.855_{-0.156}^{+0.190}$ \\
        $100$ views & $0.0009$ & $0.0051$ & $0.142_{-0.026}^{+0.029}$ & $0.838_{-0.152}^{+0.184}$ \\

        \midrule
        
        $20$ views & $0.0017$ & $0.0098$ & $0.161_{-0.026}^{+0.033}$ & $0.910_{-0.123}^{+0.144}$ \\
        $5$ views & $0.0065$ & $0.0240$ & 	$0.281_{-0.031}^{+0.037}$ & $1.037_{-0.085}^{+0.092}$ \\

        \midrule
        
        Noise $\times9$ & $0.0025$ & $0.0060$ & $0.251_{-0.045}^{+0.052}$ & $0.599_{-0.107}^{+0.130}$ \\
        Noise $\times25$ & $0.0030$ & $0.0061$ & $0.266_{-0.042}^{+0.056}$ & $0.550_{-0.095}^{+0.121}$ \\

        \midrule

        Si $\times0.6$ & $0.0155$ & $0.0287$ & $0.349_{-0.076}^{+0.090}$ & $0.644_{-0.090}^{+0.096}$\\
        Si $\times0.8$ & $0.0067$ & $0.0098$ & $0.344_{-0.060}^{+0.069}$ & $0.507_{-0.051}^{+0.057}$ \\
        Si $\times1.2$ & $0.0064$ & $0.0076$ & $0.587_{-0.062}^{+0.067}$ & $0.706_{-0.089}^{+0.102}$ \\
        Si $\times1.4$ & $0.0114$ & $0.0112$ & $0.827_{-0.078}^{+0.077}$ & $0.811_{-0.086}^{+0.093}$ \\
		\bottomrule
	\end{tabular}
	\caption{(Temperature) Sparsification summary for selected ranking metrics across selected conditions.} 
	\label{tab:uq_sparsification_T}
	%\vspace{-9pt}
\end{table}

Finally, to confirm that no single training run drives the localization signal, we recompute the full correlation $\rho(\sigma_{\rm ens}, \epsilon_{\rm mean})$ and $\operatorname{nAUSE}_\sigma$ on each of the ten leave-one-seed-out sub-ensembles (Table \ref{tab:uq_LOO}). We find that the LOO values stay within a comparably tight band, so the reported correlation values are not artifacts of any individual seed.

\begin{table}[!ht]
	\centering
    \small
    \setlength{\tabcolsep}{3.5pt}
    \renewcommand{\arraystretch}{1.5}
	\begin{tabular}{l|lll|lll} 
		\toprule
		\textbf{Condition} & $n_{\rm e}$ full $\rho$ & $n_{\rm e}$ LOO $\rho$ & $n_{\rm e}$ LOO $\operatorname{nAUSE}_\sigma$ & $T_{\rm e}$ full $\rho$ & $T_{\rm e}$ LOO $\rho$ & $T_{\rm e}$ LOO $\operatorname{nAUSE}_\sigma$ \\
        
        %\thead[l]{$n_{\rm e}$ LOO $\rho$ \\ [min, max]} & \thead[l]{$n_{\rm e}$ LOO $\operatorname{AUSE}_\sigma$ \\ [min, max]} & $T_{\rm e}$ $\rho$ & \thead[l]{$T_{\rm e}$ LOO $\rho$ \\ [min, max]} & \thead[l]{$T_{\rm e}$ LOO $\operatorname{AUSE}_\sigma$ \\ [min, max]}  \\
        \midrule
        $300$ views & $0.761$ & $[0.755, 0.767]$ & $[0.115, 0.120]$ & $0.726$ & $[0.717, 0.735]$ & $[0.138, 0.144]$\\
        $100$ views & $0.773$ & $[0.765, 0.775]$ & $[0.112, 0.128]$ & $0.732$ & $[0.726, 0.735]$ & $[0.141, 0.145]$\\

        \midrule
        
        $20$ views & $0.778$ & $[0.773, 0.783]$ & $[0.114, 0.124]$ & $0.717$ & $[0.708, 0.722]$ & $[0.160, 0.166]$\\
        $5$ views & $0.712$ & $[0.695, 0.731]$ & $[0.205, 0.240]$ & $0.611$ & $[0.599, 0.619]$ & $[0.276, 0.293]$\\

        \midrule
        
        Noise $\times9$ & $0.572$ & $[0.559, 0.568]$ & $[0.280, 0.287]$ & $0.590$ & $[0.582, 0.590]$ & $[0.251, 0.262]$\\
        Noise $\times25$ & $0.555$ & $[0.540, 0.552]$ & $[0.307, 0.324]$ & $0.601$ & $[0.585, 0.603]$ & $[0.263, 0.286]$\\

        \midrule

        Si $\times0.6$ & $0.438$ & $[0.436, 0.442]$ & $[0.405, 0.414]$ & $0.424$ & $[0.408, 0.440]$ & $[0.337, 0.356]$\\
        Si $\times0.8$ & $0.387$ & $[0.380, 0.403]$ & $[0.409, 0.422]$ & $0.448$ & $[0.437, 0.450]$ & $[0.339, 0.352]$\\
        Si $\times1.2$ & $0.166$ & $[0.158, 0.174]$ & $[0.605, 0.625]$ & $0.250$ & $[0.241, 0.256]$ & $[0.585, 0.604]$\\
        Si $\times1.4$ & $-0.017$ & $[-0.023, -0.012]$ & $[0.926, 0.951]$ & $0.096$ & $[0.088, 0.107]$ & $[0.812, 0.835]$\\
		\bottomrule
	\end{tabular}
	\caption{Leave-one-out (LOO) ranges of $\rho(\sigma_{\rm ens}, \epsilon_{\rm mean})$ and $\operatorname{nAUSE}_\sigma$ for both density and temperature.} 
	\label{tab:uq_LOO}
	%\vspace{-9pt}
\end{table}

\FloatBarrier

\subsection{Seed-Deviation Error Capture}
\label{app:error_capture}

As a complementary field-space consistency test, we project the signed ensemble error onto the vectors spanned by the ensemble-deviations. Define the ensemble-deviation subspace as $S_{\rm ens} = \operatorname{span}\{\delta_1, \dots, \delta_K \}$, where $\delta_k = \hat{m}_k - \overline{m}$ are the per-seed field deviations. Define the rank of $S_{\rm ens}$ as $r \le K-1$, since $\sum_k \delta_k = 0$ removes one degree of freedom. Then, we can define the orthogonal projection on this subspace by letting $Q\in\mathbb{R}^{M \times r}$ be an orthonormal basis of $S_{\rm ens}$, and $P_{\rm ens} = QQ^T$ as the orthogonal projection operator. This allows us to define an error-capture energy fraction

\begin{equation}
    E_{\rm cap} = \frac{||P_{\rm ens} e||^2}{||e||^2} = \frac{||Q^Te||^2}{||e||^2} \quad \in [0, 1],
    \label{eq:app_ecap}
\end{equation}

which is the fraction of squared $L^2$ norm of the signed reconstruction error that lies within the seed-deviation subspace. We additionally report enrichment relative to the rank-matched isotropic expectation $E_{\rm cap}^{\rm rand} = r / M$ for both density and temperature. Note that in our experiment, we use $K = 10$ seeds and all reported full-ensemble deviation matrices have numerical rank $r = K-1 = 9$ under a relative tolerance of $10^{-8}$ on the singular values. This is computed using the same control grid as above ($M = 4320$ per-field), so our isotropic baseline sits at around $E_{\rm cap}^{\rm rand} \approx 0.002$. Additionally, $E_{\rm cap}$ depends on the chosen control-grid discretization and Euclidean inner product. 

A high $E_{\rm cap}$ indicates directional alignment, but such alignment could be inflated by (i) reusing the same finite ensemble to construct both the signed error $e = \overline{m} - m^*$ and the deviation span, or (ii) generic spatial structure, where the signed error $e$ and the seed-deviation fields share smooth, low-frequency modes even without scene-specific alignment.

We test (i) by computing a split-seed validation test, where we split the $K = 10$ seeds into two groups $A$ and $B$. We construct the span from the per-seed field deviations in $A$, $S_A = \operatorname{span}\{\hat{m}_k - \overline{m}_A\, : \, k \in A \}$, and separately construct the signed ensemble mean error $e_B = \overline{m}_B - m^*$ from $B$. We then compute the error from ensemble $B$ captured by the span from ensemble $A$:

\begin{equation}
    E_{\rm cap}^{B\rightarrow A} = \frac{||P_Ae_B||^2}{||e_B||^2},
\end{equation}

repeated over all balanced splits, and report the mean and the $2.5$th and $97.5$th percentile interval of the finite split-sensitivity distribution. This explicitly tests whether a subspace learned from one subset of seeds captures the error of a disjoint seed subset, rather than evaluating the subspace and error from exactly the same finite ensemble. For our $K=10$ ensemble, we enumerate all $\binom{10}{5} = 252$ ordered choices of $A$ ($126$ complementary partitions evaluated in both directions), with $B$ as its complement, and report the exact finite split-sensitivity distribution, rather than a Monte Carlo approximation. We find that across matched-model conditions, our $E_{\rm cap}^{B \rightarrow A}$ values are substantially larger than the rank-matched isotropic expectation. We note that our five-member group has a numerical rank of four under the same tolerance, so its rank-matched isotropic baseline is $4/4320 \approx 0.09\%$, as opposed to the $0.2\%$ rank-9 baseline for the original $E_{\rm cap}$.

We test (ii) by using a stronger structured null: we cyclically rotate the ensemble-deviation fields $\delta_k$ relative to the signed error field $e$ through all $29$ nonzero longitude shifts, preserving spatial structure radially and latitudinally while disrupting the longitudinal alignment. More specifically, we choose a cyclic longitude shift, apply that same shift to all deviations $\delta_k$, keeping $e$ fixed. We then compute $E_{cap}$ for the shifted deviation span, and repeat for all $29$ nonzero cyclic longitude shifts. In implementation, we equivalently keep the deviation span fixed and cyclically shift the signed field error. Finally, we compute the mean $\mu$ and compare with the original $E_{cap}$ with no longitudinal rotations. Again across matched-model conditions, we find that the non-rotated $E_{cap}$ exceeds the rotation-null mean and is at the top of the distribution in all but the 100-view temperature reconstruction. The reduced capture indicates that the original longitudinal alignment contributes to the observed overlap beyond generic spatial structure alone.

\begin{table}[!ht]
	\centering
    %\small
    %\setlength{\tabcolsep}{3.5pt}
    %\renewcommand{\arraystretch}{1.5}
	\begin{tabular}{l|ccccc} 
		\toprule
		\textbf{Condition} & $E_{\rm cap}$ $\%$ & $E_{\rm cap}/ E_{\rm cap}^{\rm rand}$ (enrichment) & $E_{\rm cap}^{B\rightarrow A}$ $\%$ & rot-null $\mu$ $\%$ & $E_{\rm cap}$ rot $\%$\\
        \midrule
        noiseless ($300$v) & $7.5$ & $36\times$ & $3.6^{+2.7}_{-3.3}$ & $2.5$ & $100$ \\
        $100$ views & $4.8$ & $23\times$ & $2.5^{+2.7}_{-2.2}$ & $2.1$ & $100$ \\

        \midrule
        
        $20$ views & $27.2$ & $130\times$ & $14.4^{+8.6}_{-8.0}$ & $5.5$ & $100$ \\
        $5$ views & $32.1$ & $154\times$ & $17.9^{+11.6}_{-13.3}$ & $20.9$ & $100$\\

        \midrule
        
        Noise $\times9$ & $6.0$ & $29\times$ & $2.7^{+2.3}_{-2.0}$ & $0.9$ & $100$ \\
        Noise $\times25$ & $11.0$ & $53\times$ & $5.9^{+4.7}_{-4.4}$ & $2.3$ & $100$\\

        \midrule

        Si $\times0.6$ & $3.4$ & $16\times$ & $1.6^{+1.7}_{-1.1}$ & $2.1$ & $97$ \\
        Si $\times0.8$ & $8.0$ & $38\times$ & $4.3^{+3.8}_{-3.6}$ & $2.0$ & $100$ \\
        Si $\times1.2$ & $4.8$ & $23\times$ & $2.1^{+1.5}_{-1.5}$ & $1.7$ & $100$ \\
        Si $\times1.4$ & $1.0$ & $5\times$ & $0.6^{+0.6}_{-0.4}$ & $0.8$ & $72$ \\
		\bottomrule
	\end{tabular}
	\caption{(Density) Error-capture diagnostics as an additional field-space consistency metric.} 
	\label{tab:uq_ecap_ne}
	%\vspace{-9pt}
\end{table}

\begin{table}[!ht]
	\centering
    %\small
    %\setlength{\tabcolsep}{3.5pt}
    %\renewcommand{\arraystretch}{1.5}
	\begin{tabular}{l|ccccc} 
		\toprule
		\textbf{Condition} & $E_{\rm cap}$ $\%$ & $E_{\rm cap}/ E_{\rm cap}^{\rm rand}$ (enrichment) & $E_{\rm cap}^{B\rightarrow A}$ $\%$ & rot-null $\mu$ $\%$ & $E_{\rm cap}$ rot $\%$\\
        \midrule
        noiseless ($300$v) & $5.0$ & $24\times$ & $2.6^{+2.0}_{-1.9}$ & $1.5$ & $100$ \\
        $100$ views & $2.4$ & $11\times$ & $1.2^{+1.4}_{-0.9}$ & $1.4$ & $79$ \\

        \midrule
        
        $20$ views & $20.5$ & $98\times$ & $11.0^{+7.7}_{-7.3}$ & $4.0$ & $100$ \\
        $5$ views & $24.0$ & $115\times$ & $12.8^{+8.3}_{-8.3}$ & $18.7$ & $100$\\

        \midrule
        
        Noise $\times9$ & $9.3$ & $45\times$ & $4.6^{+3.1}_{-3.6}$ & $2.1$ & $100$ \\
        Noise $\times25$ & $20.6$ & $99\times$ & $12.8^{+7.0}_{-7.0}$ & $3.4$ & $100$\\

        \midrule

        Si $\times0.6$ & $12.1$ & $58\times$ & $6.3^{+4.4}_{-3.7}$ & $1.0$ & $100$ \\
        Si $\times0.8$ & $6.0$ & $29\times$ & $3.5^{+3.7}_{-2.8}$ & $0.8$ & $100$ \\
        Si $\times1.2$ & $2.5$ & $12\times$ & $1.4^{+1.1}_{-1.3}$ & $1.1$ & $93$ \\
        Si $\times1.4$ & $1.9$ & $9\times$ & $1.2^{+1.2}_{-1.0}$ & $0.5$ & $97$ \\
		\bottomrule
	\end{tabular}
	\caption{(Temperature) Error-capture diagnostics as an additional field-space consistency metric.} 
	\label{tab:uq_ecap_temp}
	%\vspace{-9pt}
\end{table}

Tables \ref{tab:uq_ecap_ne} and \ref{tab:uq_ecap_temp} summarize our error capture diagnostics. For selected conditions, we report the $E_{\rm cap}$ $\%$ as the fraction of the squared signed-error energy lying in the rank-9 span, and the enrichment relative to an equal-rank isotropic subspace. We also report the split-seed capture percentage $E_{\rm cap}^{B\rightarrow A}$ $\%$ with its $2.5$th-$97.5$th percentile interval, the mean capture percentage $\mu$ of the longitude-rolled deviations, and the percentile rot $\%$ of the unrotated $E_{\rm cap}$ relative to the rotation-null distribution.

\FloatBarrier
\section{Benchmark Studies and Ablations}

\subsection{Reproducibility}
\label{app:reproducibility}

Table \ref{tab:hyperparameters} provides a list of the main canonical hyperparameters used in our experiments. Note that control configurations are rerun independently within each ablation study. Because GPU hash-grid optimization uses nondeterministic operations, fixing the nominal RNG seed does not guarantee bitwise-identical training trajectories. Additionally, different studies may use different RNG streams, so numerically identical configurations can differ slightly across tables. We therefore interpret within-study mean $\pm$ standard deviation comparisons, and do not treat small cross-table differences between nominally identical controls as scientific effects.

Unless otherwise stated, we fix the model architecture to be a multiresolution hash grid with a hashmap size of $2^{20}$ $\times 2$ features, and $20$ levels spanning resolutions from $16$ (coarsest) to $512$ (finest). Separate decoder heads, each containing two hidden layers of width $64$, are used to predict log-density and log-temperature, and their biases are initialized to the midpoints of their corresponding clamp ranges, which lie within the respective LUT domains. We train with $300$ evenly-spaced views outside a $30^\circ$ held-out arc with a $3^\circ$ buffer. We use the AdamW optimizer with our asinh loss, and joint reconstructions optimize using batches of $1024$ rays for $60$k steps, while our density-only representations (Appendix \ref{app:representation_baselines}) use $30$k steps. No observation noise or abundance mismatch is applied. Our evaluation metrics are field-space errors and held-out image-space errors.

\begin{table}[!ht]
	\centering
    \renewcommand{\arraystretch}{1.5}
	\begin{tabular}{l|ll} 
		\toprule
		Group & Name & Description \\
        \midrule
        Rendering & AABB & $x_{\rm AABB} \in [-1, 1]^3$ \\
        & sample step size & $dx_{\rm AABB} = 1/256$ \\
        & LOS integration & midpoint Riemann sum \\
        \midrule
        Architecture & hash grid & $2^{20}$ hash map, $\times2$ features \\
        & levels & $20$ levels, $16 \rightarrow 512$ coarse-to-fine resolutions \\
        & MLP head & per-field, $2$ hidden layers, width $64$ \\
        \midrule
        Training & Optimizer & AdamW \\
        & learning rate & $5\times10^{-4}$ \\
        & weight decay & $10^{-5}$ \\
        & ray batch size & $b = 1024$ \\
        & steps & $s = 60$k \\
        & asinh scale & $s_c = [1.3\times10^{-2}, 4.5\times10^{-4}, 10^{-2}, 5.8\times10^{-2}]$ \\
        & ensemble seeds & \thead[l]{0xC0FFEE, 0xDECAF, 0xBADD1E, 0xD1CE \\ 0x8888, 0xB0BA, 0xFACADE, 0xCAFE \\ 0xF00D, 0xBADA55} \\
        \midrule
        Forward Model & shot coefficients & $\alpha_c = [10^{-2}, 5\times10^{-4}, 10^{-2}, 5\times10^{-2}]$ \\ 
        & noise multiplier & $\eta \in \{0, 1, 9, 25\}$ \\
        & mismatch scale & $a \in \{0.6, 0.8, 1, 1.2, 1.4\}$ \\
		\bottomrule
	\end{tabular}
	\caption{Table of canonical hyperparameters used in CoroNeRF. Note that the listed $s_c$ values are the base scales used for $\eta \le 1$. For $\eta > 1$, $s_c$ is multiplied by $\sqrt{\eta}$ as described in Appendix \ref{app:noise_view_bench}. The ordering of $\alpha_c$ and $s_c$ values corresponds to Fe XIII 1075, Fe XIII 1080, Si IX 2585, and Si IX 3935, respectively.} 
	\label{tab:hyperparameters}
	%\vspace{-9pt}
\end{table}

\subsection{Density Representation Baselines}
\label{app:representation_baselines}

Our density representational baselines below share the following properties. We only reconstruct density with the GT temperature field supplied. The observational dataset has $300$ evenly-spaced views with a $30^\circ$ arc holdout and a $3^\circ$ buffer on both ends (the arc holdout is used as validation). Each view consists of a $[-3, 3]$ R$_\odot$ FOV of two Fe XIII channels ($1075/1080$ nm), with no observational noise or abundance scale mismatch. All models take in as inputs a 3D Cartesian position mapped to the AABB. We use the AdamW optimizer, with learning rate $5\times 10^{-4}$, weight decay $10^{-5}$, and $1024$ ray batches for $30$k steps. Our loss is the fixed-scale asinh image loss. Each model is trained with $3$ seeds (0xC0FFEE, 0xDECAF, 0xBADD1E).

The first baseline we test is a classical voxel grid, which we call the \texttt{Grid} model. The optimized variable is a dense trainable log-density field with no encoder or MLP. The grid dimensions match those of the native PSI cube, with colatitude converted to latitude: $299\times142\times154$ longitude-latitude-radius, where the longitude $\in [0, 2\pi]$ is periodic (wrap-padded), latitude $\in [-\pi/2, \pi/2]$, and radius $\in [1, 30]$ R$_\odot$. Retrieving values from this trainable grid is done using trilinear interpolation via PyTorch's \texttt{grid\_sample} function, with a longitude wrap for periodicity. All grid nodes are initialized as the midpoint of the $\log n_{\rm e}$ clamp range. The number of trainable parameters in this model is the total number of voxels $6,538,532$ ($=299\cdot142\cdot154$).

We additionally test the grid representation with a first-order (Tikhonov/gradient) smoothness regularizer on the learned density grid, which we will call the \texttt{Grid-Reg} model. For a grid $g$, the regularizer has the form $\mathcal{L}_R = \lambda_{\rm lon} \langle (\Delta_{\rm lon} \, g)^2 \rangle + \lambda_{\rm lat} \langle (\Delta_{\rm lat} \, g)^2 \rangle + \lambda_{\rm r} \langle (\Delta_{\rm r} \, g)^2 \rangle$, where $\Delta$ is the first difference along each axis and $\langle \cdot \rangle$ is the mean over nodes. We set the directional weights $\lambda_{\rm lon} = \lambda_{\rm lat} = \lambda_{\rm r} = 1$. The coefficient we use for the overall total loss is $\lambda_{\rm smooth} = 10.0$, where the training objective is $\operatorname{asinhErr} + 10.0 \times \mathcal{L}_R$. We chose $10.0$ as it gave us the best preliminary results compared to other regularization values. Finally, the regularizer does not change the amount of trainable parameters ($6.5$M).

The second baseline we use is a positional encoder + MLP, which we will call the \texttt{MLP} model. The encoding frequencies are axis-wise Fourier positional encoding with $L = 6$ bands with corresponding frequencies $\{2^0, \ldots, 2^5\}$ (features are $\sin(2^k\pi \, x)$ and $\cos(2^k\pi \, x)$ per axis). This gives a total of $2$ features $\times$ $3$ axes $\times$ $6$ bands $=36$ dims. We additionally have a spherical-harmonic encoding of the unit position direction up to degree $2$ ($(2+1)^2 = 9$ dims). We concatenate the raw position with the positional encoding and spherical-harmonic encoding for a total of $48$ dims to the MLP input. Our MLP has $4$ hidden layers each with width $256$. SiLU activations connect the hidden layers, and the final layer is a linear output layer. The output is a single scalar $\log_{10} n_{\rm e}$, and the final-layer bias is initialized to the midpoint of the $\log n_{\rm e}$ clamp range. The total number of trainable parameters is $210,177$ $(= 48\cdot256 + 256 + 3\cdot(256^2+256) + (256 + 1))$.

Our final and main representation that we eventually select for the main experiments is a multiresolution hash grid as described in Appendix \ref{app:hash_grid}. The encoding hash grid has $20$ levels, $2$ features per level, with the resolutions from $16$ (coarsest) to $512$ (finest). Each level has a $2^G$ hashmap (we test $G \in \{17, 20\}$). Given a position, the hash grid produces a $40$-dimensional feature vector by concatenating the $2$ features across all $20$ levels. This is then fed into a decoder head with $2$ hidden layers each with width $64$. Activations are SiLU, and the final layer is a linear output layer. The decoder head has $6849$ trainable parameters. Our \texttt{Hash20} model has $20$ levels $\times 2^{20}$ hashmap $\times 2$ features + $6849 = 41,949,889 \approx 41.9$M parameters. Our \texttt{Hash17} model has $20$ levels $\times 2^{17}$ hashmap $\times 2$ features + $6849 = 5,249,729 \approx 5.2$M parameters.

\begin{figure}[!th]
	\begin{center}
		\includegraphics[width=1\linewidth]{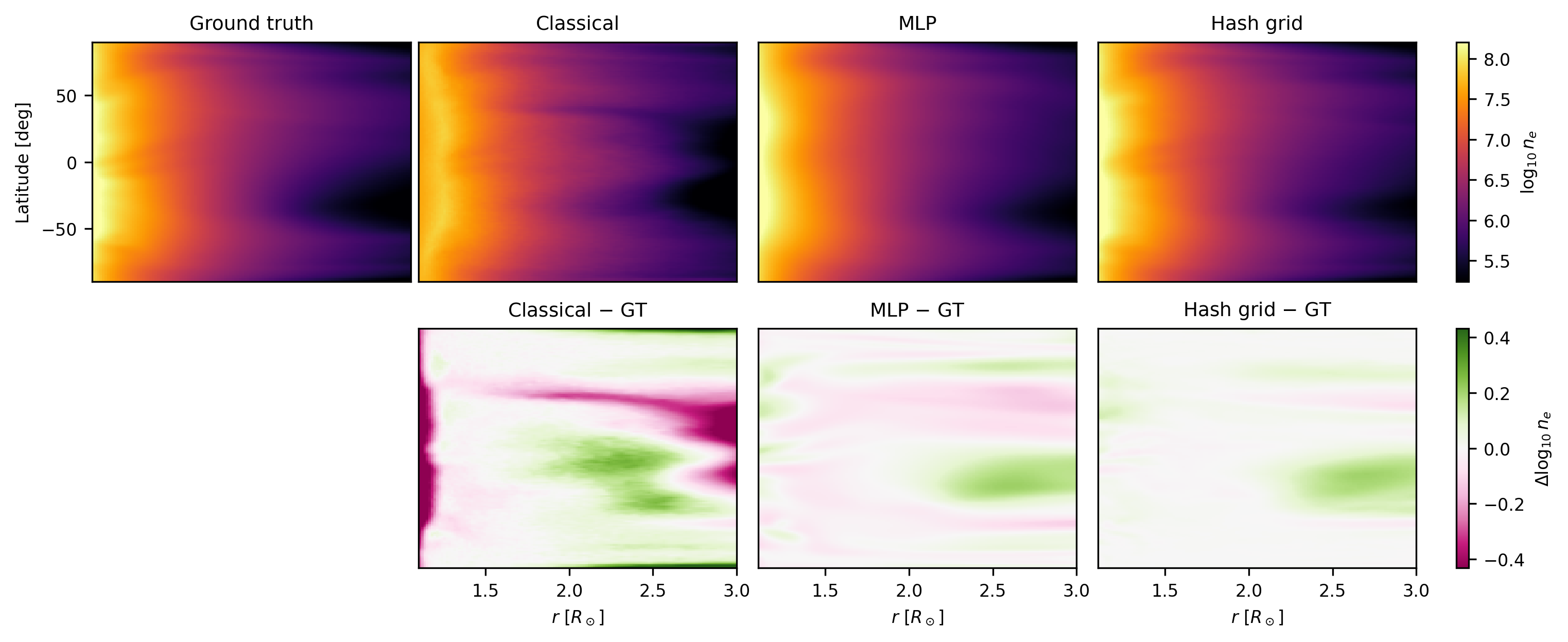}
	\end{center}
	\caption{Density-field reconstruction (with GT temperature supplied) at $0^\circ$ longitude, spanning $[1.1,3.0]$ R$_\odot$. Each panel is a meridional slice (latitude-radius plot). The top row shows the ground truth and reconstructions using classical spherical grid with Tikhonov regularization, positional-encoder + MLP, and hash grid + decoder head representations. The bottom row shows corresponding signed residuals $\Delta \log_{10} n_e = \log_{10} n_{e, \mathrm{pred}} - \log_{10} n_{e, \mathrm{GT}}$. The multiresolution hash grid is able to reconstruct consistent radial structures much farther out than the classical method, and achieves lower errors on the inner radial boundaries near the photosphere.}
	\label{fig:benchA_slice_reconstruction}
\end{figure}

\begin{table}[!ht]
	\centering
    \small
    \setlength{\tabcolsep}{3.5pt}
	\begin{tabular}{l|llll} 
		\toprule
		\textbf{Model} & $\text{MAE}_{\text{inner}} (\log_{10} n_{\rm e})$ $\downarrow$ & $\text{AbsRel}_{\text{inner}} (n_{\rm e})$ $\downarrow$ & $\text{asinhErr}(I)$ $\downarrow$ & $\text{PSNR}(I)$ $\uparrow$ \\
        \midrule
        \texttt{Hash20}  & $\mathbf{0.042 \pm <0.001}$ & $\mathbf{0.094 \pm <0.001}$ & $\mathbf{0.003 \pm <0.001}$ & $\mathbf{39.59 \pm 0.67}$ \\
        \texttt{Hash17}  & $\mathbf{0.042 \pm <0.001}$ & $0.095 \pm <0.001$ & $0.004 \pm 0.001$ & $38.77 \pm 0.65$ \\
        \midrule
        \texttt{MLP}  & $0.097 \pm 0.001$ & $0.254 \pm 0.004$ & $0.024 \pm 0.003$ & $24.65 \pm 0.62$ \\
        \midrule
        \texttt{Grid-Reg}  & $0.168 \pm <0.001$ & $0.449 \pm <0.001$ & $0.053 \pm <0.001$ & $11.57 \pm <0.01$ \\
		\texttt{Grid} & $0.217 \pm <0.001$ & $0.571 \pm <0.001$ & $0.080 \pm <0.001$ & $11.30 \pm <0.01$ \\
		\bottomrule
	\end{tabular}
	\caption{Density representation benchmark. Each row indicates a model configuration, and metrics are reported as mean $\pm$ standard deviation over 3 seeds, with deviations below $5\times10^{-4}$ reported as $<0.001$. The first column indicates the experiment family, and the last four columns are the relevant metrics. Image metrics are averaged over views and channels, and density metrics are averaged over inner coronal radial shells $1.1$-$2$ R$_\odot$. The multiresolution hash grid outperforms the MLP by more than $2\times$, and the classical grid by about $4\times$.} 
	\label{tab:benchA_compare}
	\vspace{-9pt}
\end{table}

Figure \ref{fig:benchA_slice_reconstruction} shows a meridional (latitude-radius) slice of the reconstructed density at $0^\circ$ longitude for various representational baselines, providing a complementary view of the reconstructed corona compared to a latitude-longitude shell shown in Figure \ref{fig:benchA_reconstruction} in the main text. Table \ref{tab:benchA_compare} provides the field-space and image-space metrics for our density representations. We find that the two tested hash-table sizes, $G \in \{17, 20\}$, achieve nearly identical field error.

\FloatBarrier

\subsection{Loss Ablation Study}
\label{app:loss_ablation}

In our loss ablation study, we keep the same model architecture and observational setup as those described in Appendix \ref{app:reproducibility}, except we introduce observational noise at $\eta = 1$ and alter the objective function. We use the following objective family:

\begin{equation}
    \mathcal{L}_{\mathrm{train}} = \lambda_G \mathcal{L}_G + \lambda_A \mathcal{L}_A,
\end{equation}

where $\mathcal{L}_G$ is the heteroscedastic Gaussian image loss, and $\mathcal{L}_A$ is the fixed-scale asinh image loss. More specifically, given ray $q$ and channel $c$, the heteroscedastic Gaussian term is the inverse-variance-weighted squared image residual $((\hat{I}_{\theta, q, c} - y_{q, c})/\sigma_{q,c})^2$, which is equivalent for optimization to the Gaussian NLL up to a positive constant factor and additive terms independent of $\theta$. The whitening-scale $\sigma_{q,c}$ is the target-$\sigma$, or the known per-pixel, per-channel observation noise standard deviation of our generative model (Equation \ref{eq:noise_model}): this is the exact heteroscedastic standard deviation used to add noise to the training images. By contrast, the asinh term uses $| \operatorname{asinh}(\hat{I}_{q,c}/s_c) - \operatorname{asinh}(y_{q,c}/s_c) |$, where $s_c$ is a channel-dependent fixed characteristic intensity scale. Our experiment families vary $\lambda_G$ and $\lambda_A$ so as to test objectives that are pure Gaussian, mixed Gaussian and asinh, and pure asinh. Each experiment family is trained with 3 seeds, and replicate runs use different observation-noise realizations coupled to the run seed.

\begin{figure}[!th]
	\begin{center}
		\includegraphics[width=1\linewidth]{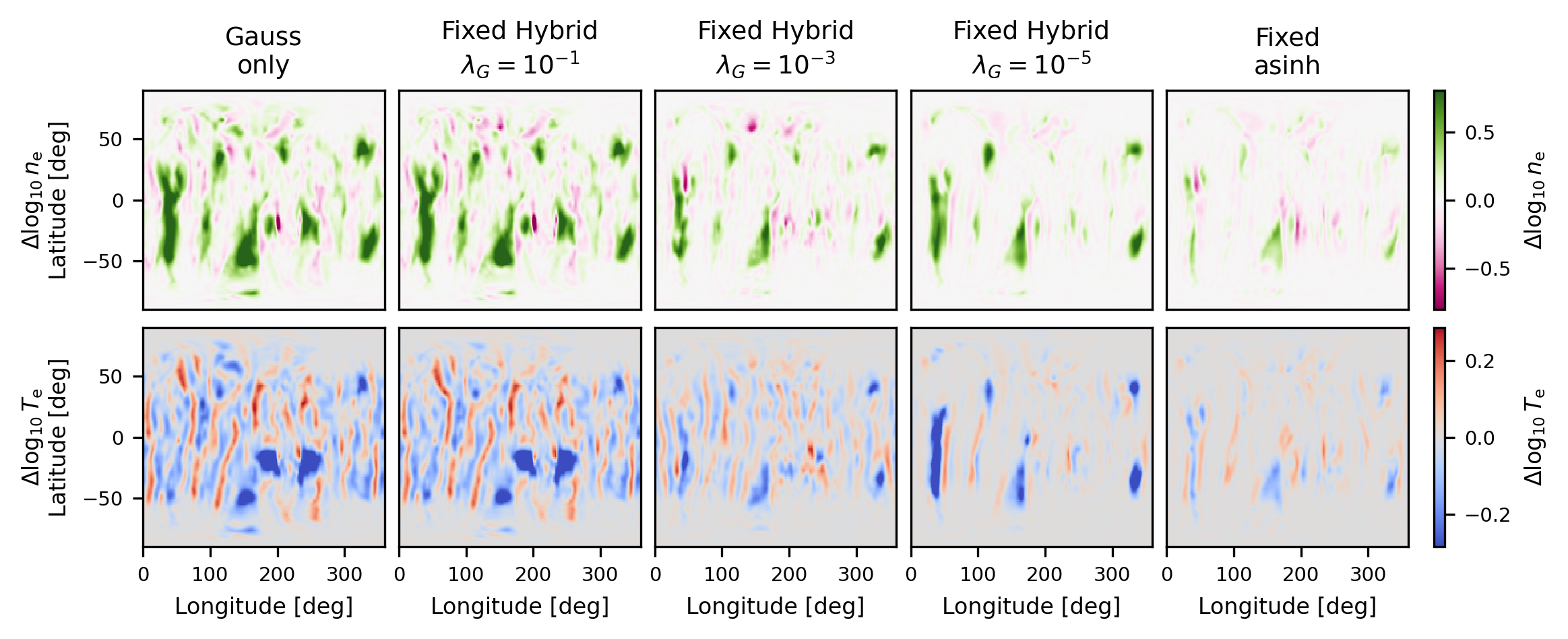}
	\end{center}
	\caption{Density and temperature reconstructed residuals for different loss metric settings. We find that the Gaussian-only loss performs worst while the asinh-only loss performs best, with hybrid losses landing in between the two regimes.}
	\label{fig:benchE_loss_shell}
\end{figure}

\begin{table}[!ht]
	\centering
	\begin{tabular}{l|lll} 
		\toprule
		\textbf{Objective} & $\text{MAE}_{\text{inner}} (\log_{10} n_{\rm e})$ $\downarrow$ & $\text{MAE}_{\text{inner}} (\log_{10} T_{\rm e})$ $\downarrow$ & $\text{asinhErr}(I)$ $\downarrow$ \\
        \midrule
        Gaussian-only $\lambda_G = 1$ & $0.145 \pm 0.010$ & $0.075 \pm 0.020$ & $0.060 \pm 0.006$ \\
        Fixed hybrid $\lambda_G = 10^{-1}$ & $0.134 \pm 0.009$ & $0.064 \pm 0.014$ & $0.047 \pm 0.005$ \\
        Fixed hybrid $\lambda_G = 10^{-3}$ & $0.087 \pm 0.001$ & $0.036 \pm 0.001$ & $0.032 \pm 0.002$ \\
        Fixed hybrid $\lambda_G = 10^{-5}$ & $0.061 \pm 0.002$ & $0.022 \pm 0.002$ & $\mathbf{0.021 \pm <0.001}$ \\
        Fixed asinh only $\lambda_G = 0$ & $\mathbf{0.047 \pm 0.001}$ & $0.017 \pm <0.001$ & $0.022 \pm <0.001$ \\
        \midrule
        Target-$\sigma$ asinh $\lambda_G = 0$ & $0.047 \pm <0.001$ & $\mathbf{0.016 \pm <0.001}$ & $0.022 \pm 0.001$ \\
		\bottomrule
	\end{tabular}
	\caption{Loss ablation benchmark. For the Gaussian-only experiment, $\lambda_A = 0$, otherwise it is fixed at $\lambda_A = 1$. We find that objectives that favor the asinh term $\mathcal{L}_A$ generally perform better than those that favor the heteroscedastic Gaussian loss term.} 
	\label{tab:benchE_compare}
	\vspace{-9pt}
\end{table}

Under the tested training protocol, the pure asinh objective yields lower log-field $\operatorname{MAE}$ than the Gaussian-only objective (Figure \ref{fig:benchE_loss_shell} and Table \ref{tab:benchE_compare}). We therefore use it as the task-aware objective for the subsequent reconstruction experiments. As an auxiliary oracle variant, the final row, our target-$\sigma$ case, replaces $s_c$ in $\mathcal{L}_A$ with the synthetic generator's per-pixel target $\sigma$. The corresponding loss term becomes $| \operatorname{asinh}(\hat{I}_{q,c}/\sigma_{q,c}) - \operatorname{asinh}(y_{q,c}/\sigma_{q,c}) |$, and is thus included only as an oracle diagnostic ablation. Finally, we note that this ablation compares complete objective choices rather than separately isolating the effects of intensity transformation, residual norm, and weighting.

\subsection{Compute and Runtime}
\label{app:compute_ablation}

% GPU usage
Our multiresolution hash grid has around $40$M parameters ($2M$ per hash table, $20$ levels), and is trained in a cluster GPU setting: training a single reconstruction takes $\approx3$ GPU-hours on an $80$ GB NVIDIA A$100$, and so a $K = 10$ ensemble takes $\approx30$ aggregate GPU-hours. The ten localization conditions reported in Tables \ref{tab:uq_correlations_ne}-\ref{tab:uq_ecap_temp} require approximately $300$ aggregate GPU-hours. Rendering a $300$-view set, as a dataset or post-training analysis, can take $\approx10$-$100$ minutes on a local NVIDIA RTX 4090 depending on the resolution and integration step size. Finally, diagnostics such as $\sigma_{\rm ens}$ are cheap as they only require $K$ forward evaluations.

We performed pilot tests to pick batch size $b$ and step count $s$ from a compute-quality trade-off ablation study (Table \ref{tab:data_ablation_data_compute}). Note that these pilot runs use the earlier density-only configuration with no observational noise, and are not directly comparable to the noise-view sweep of the joint thermodynamic reconstruction in Figure \ref{fig:benchC_heatmaps}. We find that field error varies modestly across the tested settings, and note that because these tests were done only for hyperparameter selection at an earlier stage of the pipeline, they should not be used for scientific conclusions.  

\begin{table}[!ht]
	\centering
	\begin{tabular}{l|llll|lll} 
		\toprule
		\textbf{Exp} & $b$ & $s$ & $V$ & $D$ & \textbf{MAE}($\log_{10} n_e$) $\downarrow$ & \textbf{PSNR}(I) $\uparrow$ & \textbf{MSE}($\log_{10} I$) $\downarrow$ \\
		\midrule
		big  & $4096$ & $60$k & $30$ & $64$ & $0.096$ & $30.56$ & $1.75 \times 10^{-4}$ \\
		big  & $4096$ & $60$k & $1000$ & $256$ & $0.097$ & $30.74$ & $1.78 \times 10^{-4}$ \\
		\midrule
		med  & $4096$ & $30$k & $1000$ & $256$ & $0.106$ & $26.60$ & $5.46 \times 10^{-4}$ \\
		med  & $4096$ & $30$k & $30$ & $64$ & $0.107$ & $27.47$ & $4.81 \times 10^{-4}$ \\
		\midrule
		base  & $1024$ & $30$k & $1000$ & $256$ & $0.109$ & $24.88$ & $6.40 \times 10^{-4}$ \\
		base  & $1024$ & $30$k & $30$ & $64$ & $0.109$ & $27.82$ & $5.01 \times 10^{-4}$ \\
		\bottomrule
	\end{tabular}
	\caption{Compute-quality trade-off ablation study. We vary the batch size $b$, steps $s$, views $V$, and image dimension $D$. We find that the log-density $\operatorname{MAE}$ varies modestly across the tested settings, with slightly better performance at $b = 4096$, $s = 60$k.} 
	\label{tab:data_ablation_data_compute}
	%\vspace{-9pt}
\end{table}

\FloatBarrier
\subsection{Spectral Channel Ablation Study}
\label{app:spectral_bench}
% full bench B2 results

This section provides details on our spectral channel ablation study. We ablate the number of spectral channels for a total of 5 experiment families: Fe XIII $1075$, Si IX $3935$, Fe XIII pair, Si IX pair, and All four. Each experiment family is trained with 3 seeds.

\begin{table}[!ht]
	\centering
    %\small
    %\footnotesize
    %\setlength{\tabcolsep}{3.5pt}
	\begin{tabular}{l|lll} 
		\toprule
		\textbf{Channel Set} & $\text{MAE}_{\text{inner}} (\log_{10} n_{\rm e})$ $\downarrow$ & $\text{MAE}_{\text{inner}} (\log_{10} T_{\rm e})$ $\downarrow$ & $\text{asinhErr}$(observed channels) $\downarrow$ \\
        \midrule
        all4  & $\mathbf{0.039 \pm <0.001}$ &  $\mathbf{0.014 \pm <0.001}$  & $0.008 \pm 0.002$ \\
        \midrule
        Fe XIII pair  & $0.097 \pm 0.004$ &  $0.023 \pm 0.001$  & $0.009 \pm 0.003$ \\
        Si IX pair  & $0.068 \pm 0.001$  & $0.058 \pm 0.001$ & $\mathbf{0.004 \pm 0.001}$ \\
        \midrule
        Fe $1075$  & $0.605 \pm 0.004$  & $0.072 \pm <0.001$  & $0.006 \pm 0.001$ \\
		Si $3935$ & $0.530 \pm 0.028$ & $0.195 \pm 0.006$ & $0.005 \pm 0.002$ \\
		\bottomrule
	\end{tabular}
	\caption{Reconstruction metrics for our spectral ablation study. The four-line configuration achieves the lowest field errors and low across-seed variability in this three-seed ablation. We can also observe that a good image-space reconstruction does not imply a good physical-field reconstruction: while the Si IX pair has the lowest image-space validation loss, it still has higher physical-field errors.} 
	\label{tab:benchB_compare}
	\vspace{-9pt}
\end{table}

\begin{figure}[!th]
	\begin{center}
		\includegraphics[width=1\linewidth]{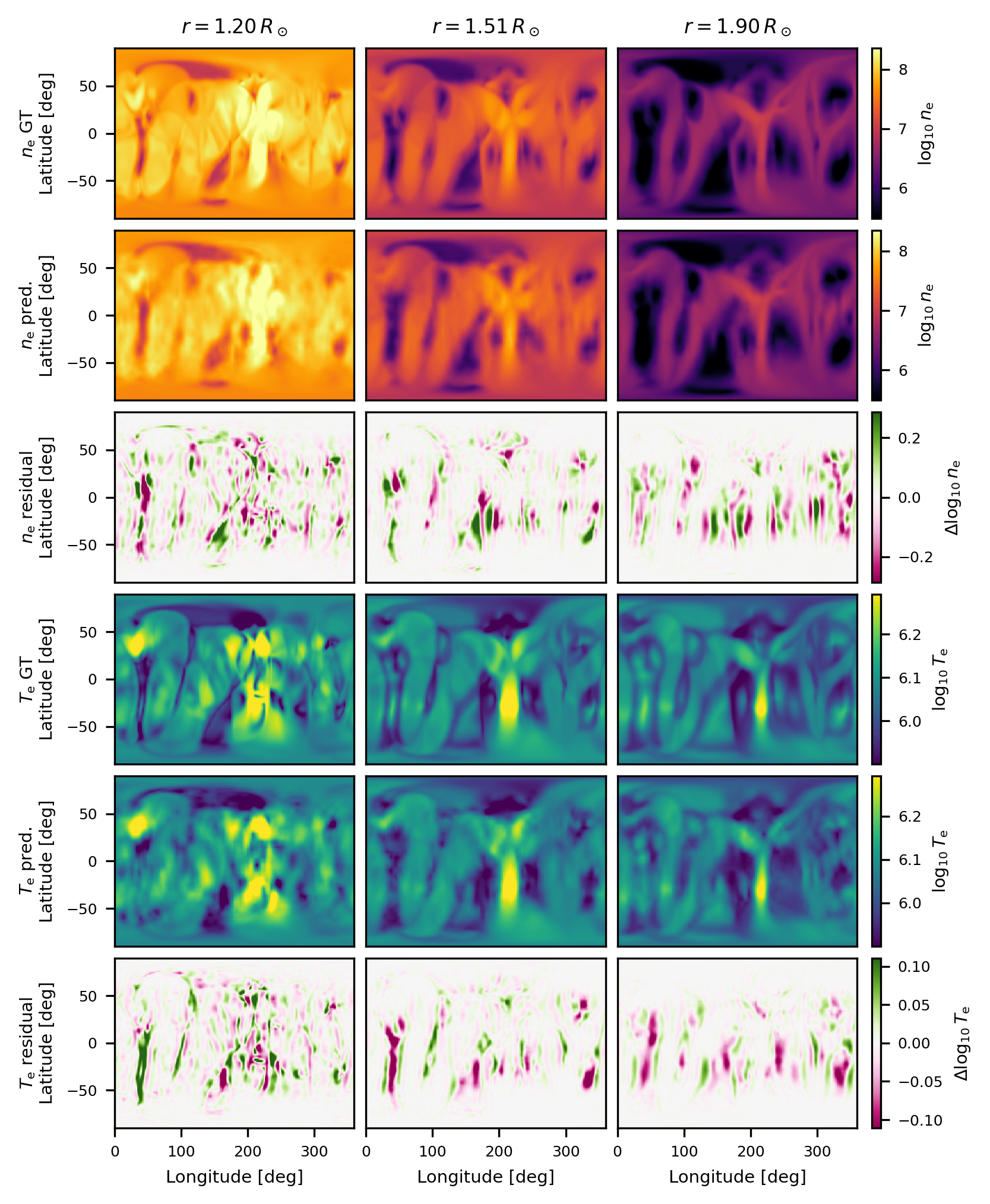}
	\end{center}
	\caption{Field Reconstruction at three selected radii: $1.20, 1.51, 1.90$ R$_\odot$ (columns). We show the field-space GT, prediction, and signed residuals $\Delta m = \hat{m} - m^*$ for $m \in \{\log_{10} n_{\rm e}, \log_{10} T_{\rm e} \}$.}
	\label{fig:benchB_multi_radius_shell}
\end{figure}

\begin{figure}[!th]
	\begin{center}
		\includegraphics[width=1\linewidth]{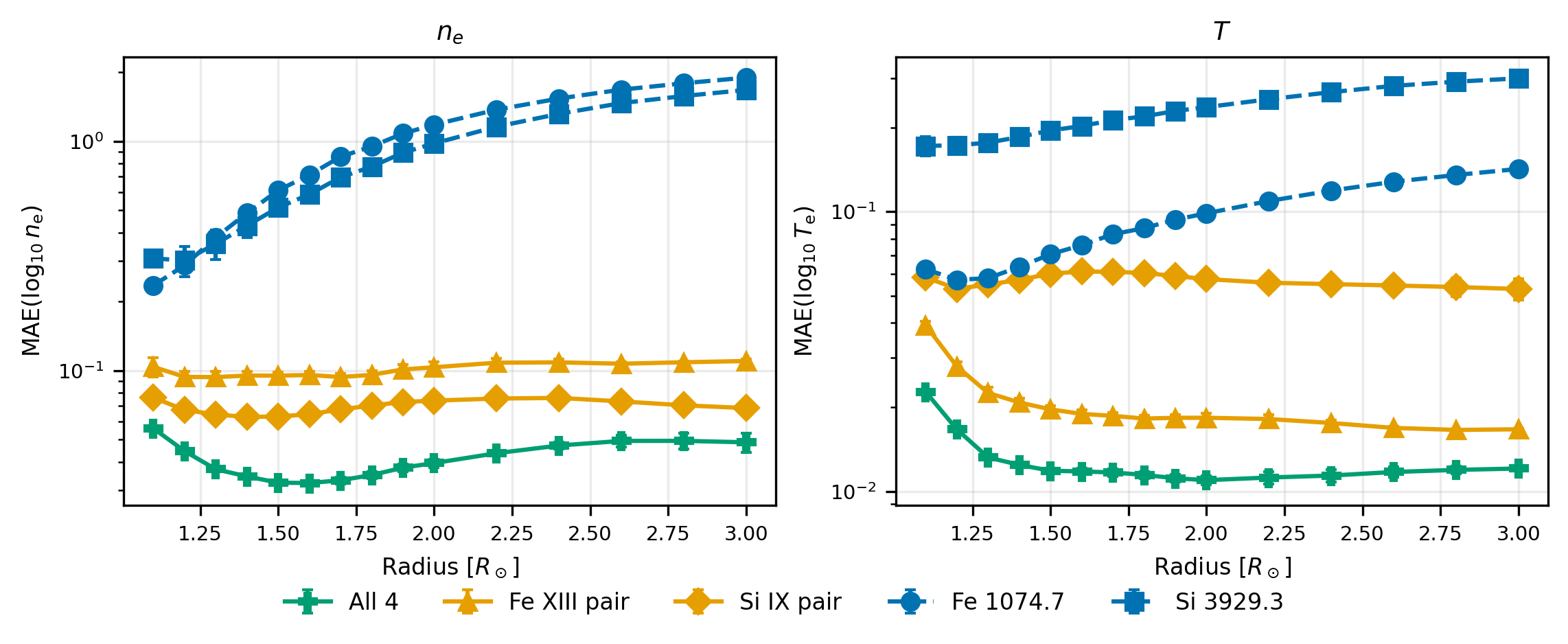}
	\end{center}
	\caption{Field-space $\operatorname{MAE}$ radial dependence for the spectral ablation study. Color denotes the number of spectral lines, and marker shape denotes the line set. We find that having at least a pair of lines substantially reduces radial errors across $1.1$-$3$ R$_\odot$.}
	\label{fig:benchB_radial_curves}
\end{figure}

\begin{figure}[!th]
	\begin{center}
		\includegraphics[width=1\linewidth]{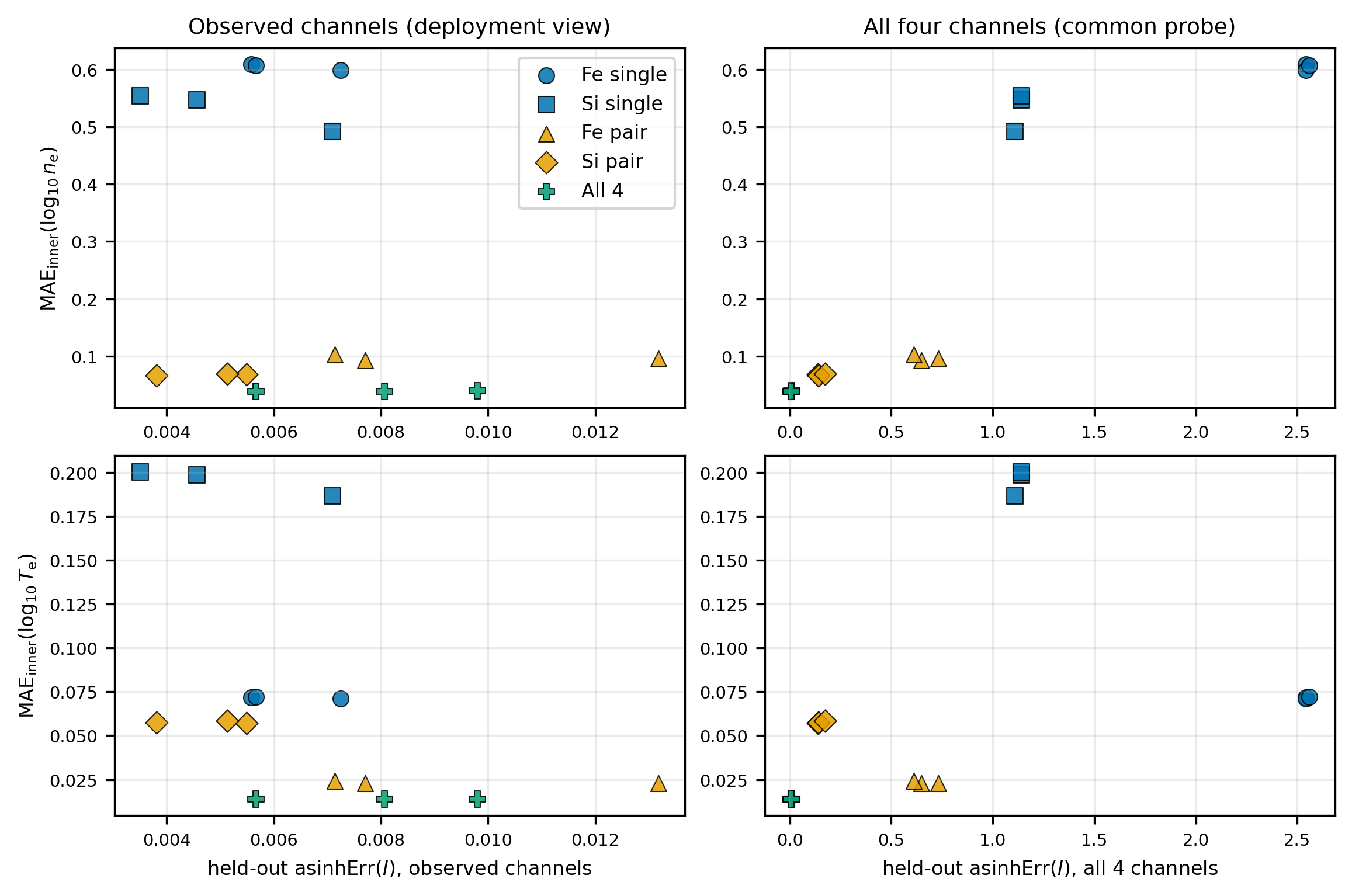}
	\end{center}
	\caption{3D field-space $\operatorname{MAE}$ vs 2D held-out image-space validation loss, showing that a good image-space reconstruction need not imply a good field-space reconstruction. Each point is one seed from the spectral ablation benchmark trained with different sets of spectral channels. We find that fewer-channel models can achieve lower observed-channel image error than the four-line model despite worse field recovery.}
	\label{fig:benchB_scatter_appendix}
\end{figure}

Table \ref{tab:benchB_compare} summarizes the field-space and image-space performance for all five experiment families, and Figure \ref{fig:benchB_multi_radius_shell} shows a sample joint reconstruction at three selected radii and corresponding residual panels. We also provide radial profiles of the field-space $\operatorname{MAE}$s (Figure \ref{fig:benchB_radial_curves}). Finally, Figure \ref{fig:benchB_scatter_appendix} shows image-space versus field-space error for all five experiment families, for both density and temperature (as opposed to the density-only Figure \ref{fig:benchB_scatter} in the main text). We note that for visualizations of radial shells, we select the median-performing model (out of three seeds) using $\operatorname{MAE}$ of the inner band temperature field.

\FloatBarrier
\subsection{Noise/View Ablation Study}
\label{app:noise_view_bench}

This section details the noise-view stress ablation study described in Figure \ref{fig:benchC_heatmaps} in the main text. We keep the same model architecture and observational setup as described in Appendix \ref{app:reproducibility}. However, we ablate the number of views $V$ by selecting evenly sampled views from the original $300$ views. Likewise, we also ablate the noise scale $\eta$. Additionally, we calibrate the per-channel asinh scale to the noise level: $s_c$ is fixed at the baseline for $\eta \in \{0, 1\}$, and scaled by $\sqrt{\eta}$ for higher noise, so it tracks the noise standard deviation (variance scales as $\eta$, standard deviation as $\sqrt{\eta}$). The scale is constant within any single run.

Our ablation study has four noise levels ($\eta \in \{0, 1, 9, 25\}$) crossed with four view counts ($V \in \{5, 20, 100, 300\}$), for a total of 16 distinct combinations trained with three different seeds for 48 total models. For noisy conditions, the three replicate seeds also use distinct observation-noise realizations derived from the run seed.

\begin{figure}[!th]
	\begin{center}
		\includegraphics[width=1\linewidth]{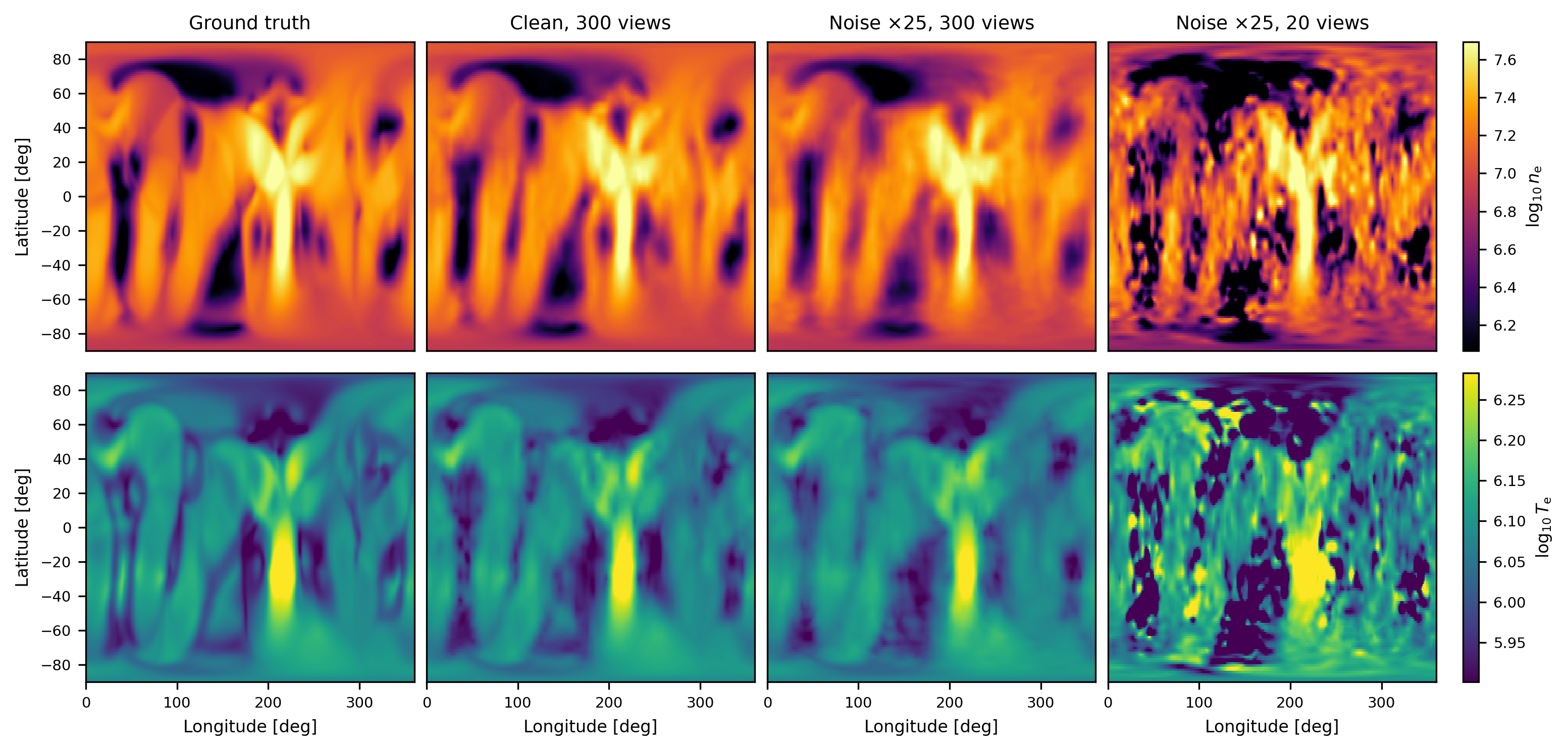}
	\end{center}
	\caption{Robustness in physical field reconstruction. The columns show ground truth and the reconstructed density (top row) and temperature (bottom row) for different combinations of views and noise multipliers. We see that with 300 views, there is moderate degradation between a clean reconstruction with no noise and one with a noise multiplier of 25, while retaining low absolute error. However, once we reduce the number of views significantly to 20 views (last column), the combined effect of a large noise multiplier and sparse views is a failure mode for physical field reconstruction.}
	\label{fig:benchC_field_reconstruction}
\end{figure}

\begin{table}[!ht]
	\centering
	\begin{tabular}{l|lll} 
		\toprule
		\textbf{Noise Setting} & $\text{MAE}_{\text{inner}} (\log_{10} n_{\rm e})$ $\downarrow$ & $\text{MAE}_{\text{inner}} (\log_{10} T_{\rm e})$ $\downarrow$ & $\text{asinhErr}(I)$ $\downarrow$ \\
        \midrule
        $\eta=0$,\ 5 views   & $0.410 \pm 0.020$      & $0.070 \pm 0.005$      & $1.454 \pm 0.041$ \\
        $\eta=0$,\ 20 views    & $0.122 \pm 0.023$      & $0.026 \pm 0.003$      & $0.458 \pm 0.060$ \\
        $\eta=0$, 100 views    & $0.039 \pm {<}0.001$   & $0.014 \pm {<}0.001$   & $0.007 \pm 0.001$ \\
        $\eta=0$, 300 views    & $0.039 \pm {<}0.001$   & $0.014 \pm {<}0.001$   & $0.007 \pm 0.001$ \\
        \midrule
        $\eta=1$,\ 5 views   & $0.516 \pm 0.026$      & $0.126 \pm 0.010$      & $1.513 \pm 0.046$ \\
        $\eta=1$,\ 20 views    & $0.209 \pm 0.018$      & $0.044 \pm 0.003$      & $0.823 \pm 0.050$ \\
        $\eta=1$, 100 views    & $0.052 \pm 0.002$      & $0.018 \pm {<}0.001$   & $0.039 \pm {<}0.001$ \\
        $\eta=1$, 300 views    & $0.047 \pm 0.001$      & $0.017 \pm {<}0.001$   & $0.022 \pm {<}0.001$ \\
        \midrule
        $\eta=9$,\ 5 views   & $0.519 \pm 0.021$      & $0.970 \pm 0.370$      & $1.565 \pm 0.050$ \\
        $\eta=9$,\ 20 views    & $0.275 \pm 0.037$      & $0.115 \pm 0.030$      & $1.239 \pm 0.084$ \\
        $\eta=9$, 100 views    & $0.069 \pm 0.003$      & $0.022 \pm {<}0.001$   & $0.092 \pm 0.004$ \\
        $\eta=9$, 300 views    & $0.057 \pm 0.002$      & $0.019 \pm {<}0.001$   & $0.041 \pm 0.001$ \\
        \midrule
        $\eta=25$,\ 5 views  & $0.760 \pm 0.458$      & $2.404 \pm 1.154$      & $1.529 \pm 0.059$ \\
        $\eta=25$,\ 20 views   & $0.336 \pm 0.077$      & $0.380 \pm 0.138$      & $1.418 \pm 0.107$ \\
        $\eta=25$, 100 views   & $0.081 \pm 0.005$      & $0.026 \pm 0.001$      & $0.136 \pm 0.008$ \\
        $\eta=25$, 300 views   & $0.063 \pm 0.002$      & $0.021 \pm {<}0.001$   & $0.054 \pm 0.001$ \\
        \bottomrule
	\end{tabular}
	\caption{Noise-view stress benchmark results (also shown in Figure \ref{fig:benchC_heatmaps} in the main text). With sufficient angular coverage, the reconstruction remains robust even under strong heteroscedastic observation noise: increasing the noise multiplier to $\times25$ at 300 training views only moderately increases the density and temperature field errors. In contrast, sparse-view regimes are substantially less stable, and the combination of high noise and sparse views produces a clear failure mode in latent physical-field recovery. The training objective uses the noise-matched $s_c$ scaling described above, while held-out $\text{asinhErr}$ is evaluated on noise-free held-out views using the same base per-channel scales across all $\eta$.} 
	\label{tab:benchC_compare}
	\vspace{-9pt}
\end{table}

\begin{figure}[!th]
	\begin{center}
		\includegraphics[width=1\linewidth]{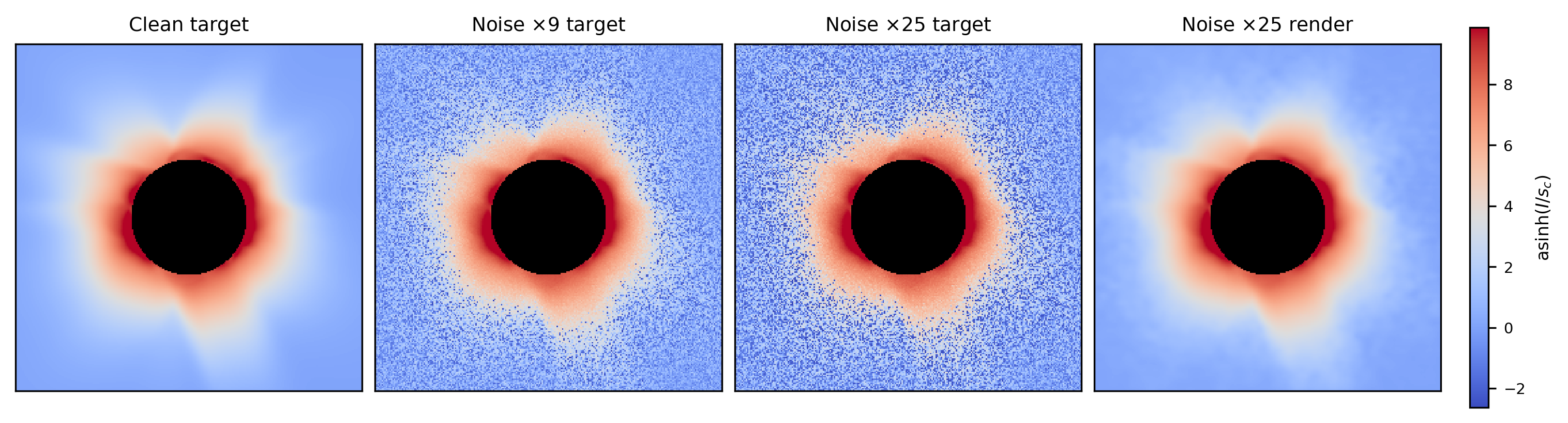}
	\end{center}
	\caption{Demonstration of the accompanying denoising effect in the framework. We show the clean ground truth target, and sample noisy training observations. The final column shows a reconstructed render using a model trained on noise $\times25$ images, where multiview consistency suppresses pixel-scale noise in the reconstructed render.}
	\label{fig:benchC_obs_reconstruction}
\end{figure}

Figure \ref{fig:benchC_field_reconstruction} shows a sample reconstruction at $1.51$ R$_\odot$, and Table \ref{tab:benchC_compare} shows detailed results of our ablation study. Additionally, we show an apparent denoising effect of training on noisy observations with sufficient views in Figure \ref{fig:benchC_obs_reconstruction}.

\FloatBarrier
\subsection{Abundance Mismatch Ablation Study}
% full bench D results
\label{app:abundance_bench}

\begin{figure}[!th]
	\begin{center}
		\includegraphics[width=1\linewidth]{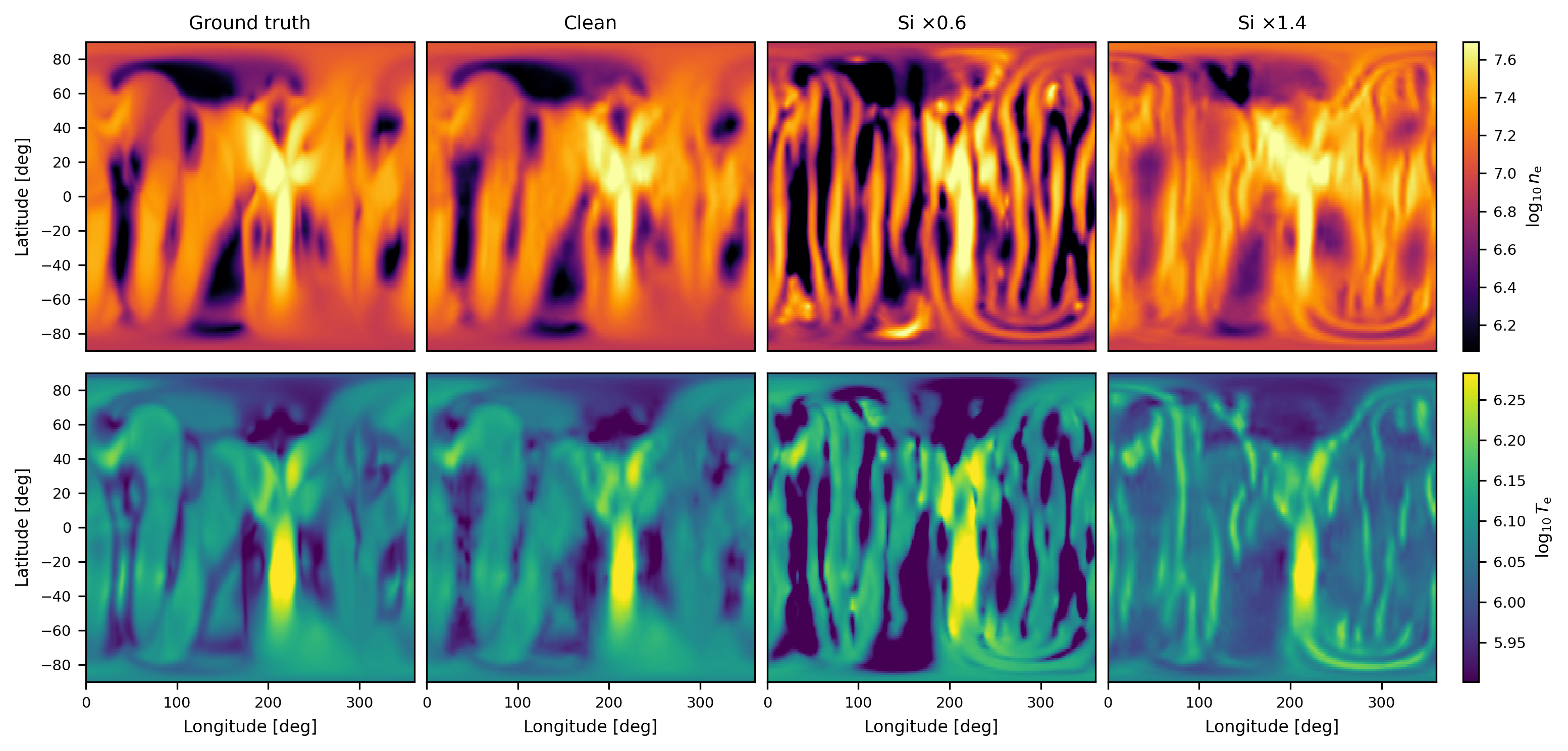}
	\end{center}
	\caption{Errors in physical field reconstruction due to forward model mismatch. The columns show ground truth and the reconstructed density (top row) and temperature (bottom row) for different abundance mismatch scales. We see a clear failure mode when the Si IX intensities are scaled by $\pm 40\%$: the reconstructed field has spurious sharp features that do not exist in the ground truth fields.}
	\label{fig:benchD_field_reconstruction}
\end{figure}

This section details an abundance scale ablation study that probes a controlled common-mode forward-model mismatch. We keep the same model architecture and observational setup as described in Appendix \ref{app:reproducibility}. However, we ablate the abundance scale $a$ of the Si IX channels to produce a forward model mismatch between the model used to generate synthetic observations and that used to train CoroNeRF. Because this ablation has no added observation noise, $I_c^{\rm obs} = I_c^{\rm true} = aI_c^{\rm raw}$ for the Si IX channels with varying $a$ while the Fe XIII channels remain unscaled; the inversion renderer continues to use the nominal $a = 1$ for all channels. Our study has 5 experiment families that vary $a \in \{0.6, 0.8, 1.0, 1.2, 1.4\}$, each trained with 3 seeds for a total of 15 runs. Figure \ref{fig:benchD_field_reconstruction} shows sample reconstructions at $1.51$ R$_\odot$, and Table \ref{tab:benchD_abundance_mismatch} details the field-space and image-space errors.

\begin{table}[!ht]
	\centering
    \small
    \setlength{\tabcolsep}{3.0pt}
	\begin{tabular}{l|lllll} 
		\toprule
		\textbf{$a$} & $\text{asinhErr}(I)$ $\downarrow$ & $n_{\rm e}$ $\operatorname{MAE}$ $\downarrow$ & $n_{\rm e}$ $\operatorname{ME}$ & $T_{\rm e}$ $\operatorname{MAE}$ $\downarrow$ & $T_{\rm e}$ $\operatorname{ME}$ \\
        \midrule
        $0.6$ & $0.030 \pm <0.001$ & $0.301 \pm 0.006$ & $-0.211 \pm 0.004$ & $0.076 \pm 0.001$ & $-0.030 \pm 0.002$ \\
        $0.8$ & $0.017 \pm <0.001$ & $0.146 \pm 0.005$ & $-0.076 \pm 0.004$ & $0.036 \pm 0.001$ & $-0.010 \pm 0.001$ \\
        $1.0$ & $\mathbf{0.006 \pm <0.001}$ & $\mathbf{0.039 \pm <0.001}$ & $-0.001 \pm 0.001$ & $\mathbf{0.014 \pm <0.001}$ & $\mathbf{0.000 \pm <0.001}$ \\
        $1.2$ & $0.016 \pm <0.001$ & $0.092 \pm 0.001$ & $0.062 \pm 0.001$ & $0.023 \pm <0.001$ & $-0.002 \pm <0.001$ \\
        $1.4$ & $0.027 \pm <0.001$ & $0.142 \pm 0.001$ & $0.107 \pm 0.001$ & $0.031 \pm <0.001$ & $-0.006 \pm <0.001$ \\
		\bottomrule
	\end{tabular}
	\caption{Abundance-mismatch benchmark. We scale the Si IX intensity channels in the synthetic observations by $a$ while evaluating against the clean ground-truth physical fields. $\operatorname{MAE}$ reports inner-band absolute log-field error, while ME reports signed inner-band errors. The clean setting, corresponding to scale $1.0$, gives the lowest image and field errors. Scaling the Si IX channels below or above the correct value produces systematic signed density errors, indicating that forward-model mismatch can be absorbed into incorrect latent physical fields.} 
	\label{tab:benchD_abundance_mismatch}
	\vspace{-9pt}
\end{table}

\FloatBarrier
\subsection{Rendered Views}

\begin{figure}[!th]
	\begin{center}
		\includegraphics[width=1\linewidth]{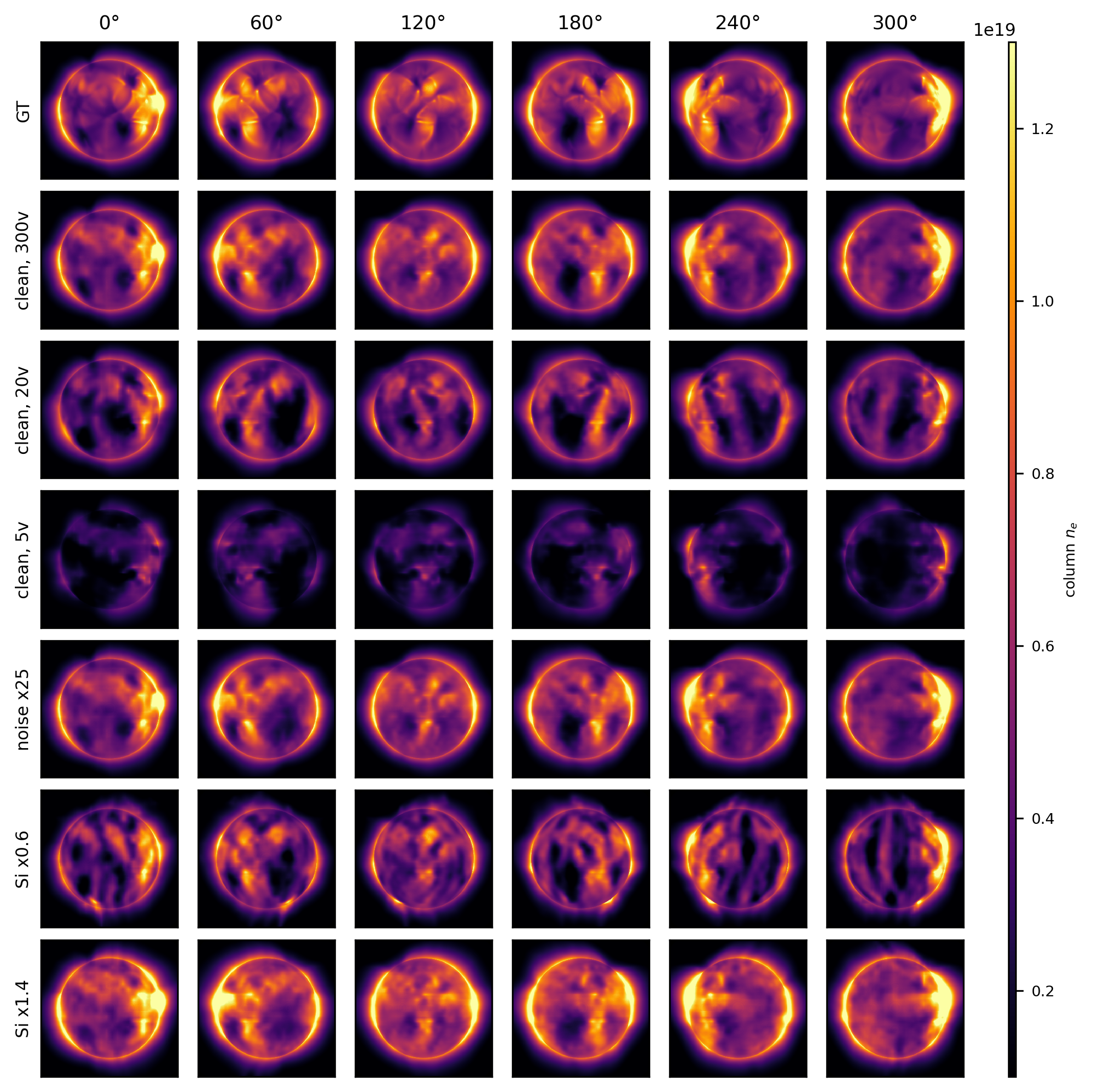}
	\end{center}
	\caption{LOS projections of recovered column density for selected conditions and view angles.}
	\label{fig:benchF_filmstrip_density}
\end{figure}

\begin{figure}[!th]
	\begin{center}
		\includegraphics[width=1\linewidth]{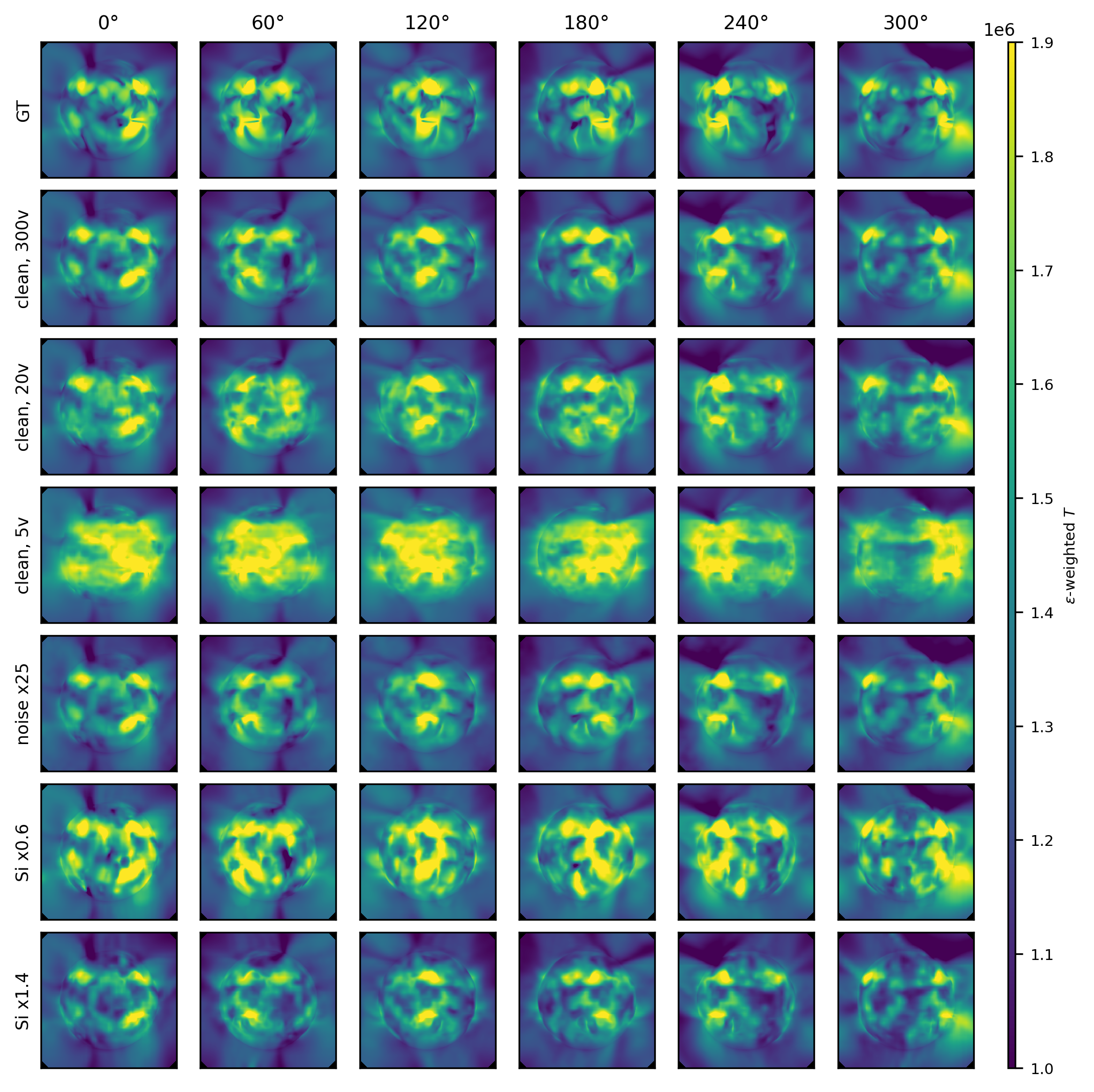}
	\end{center}
	\caption{LOS projections of recovered emissivity-weighted temperature for selected conditions and view angles.}
	\label{fig:benchF_filmstrip_temp}
\end{figure}

\begin{figure}[!th]
	\begin{center}
		\includegraphics[width=1\linewidth]{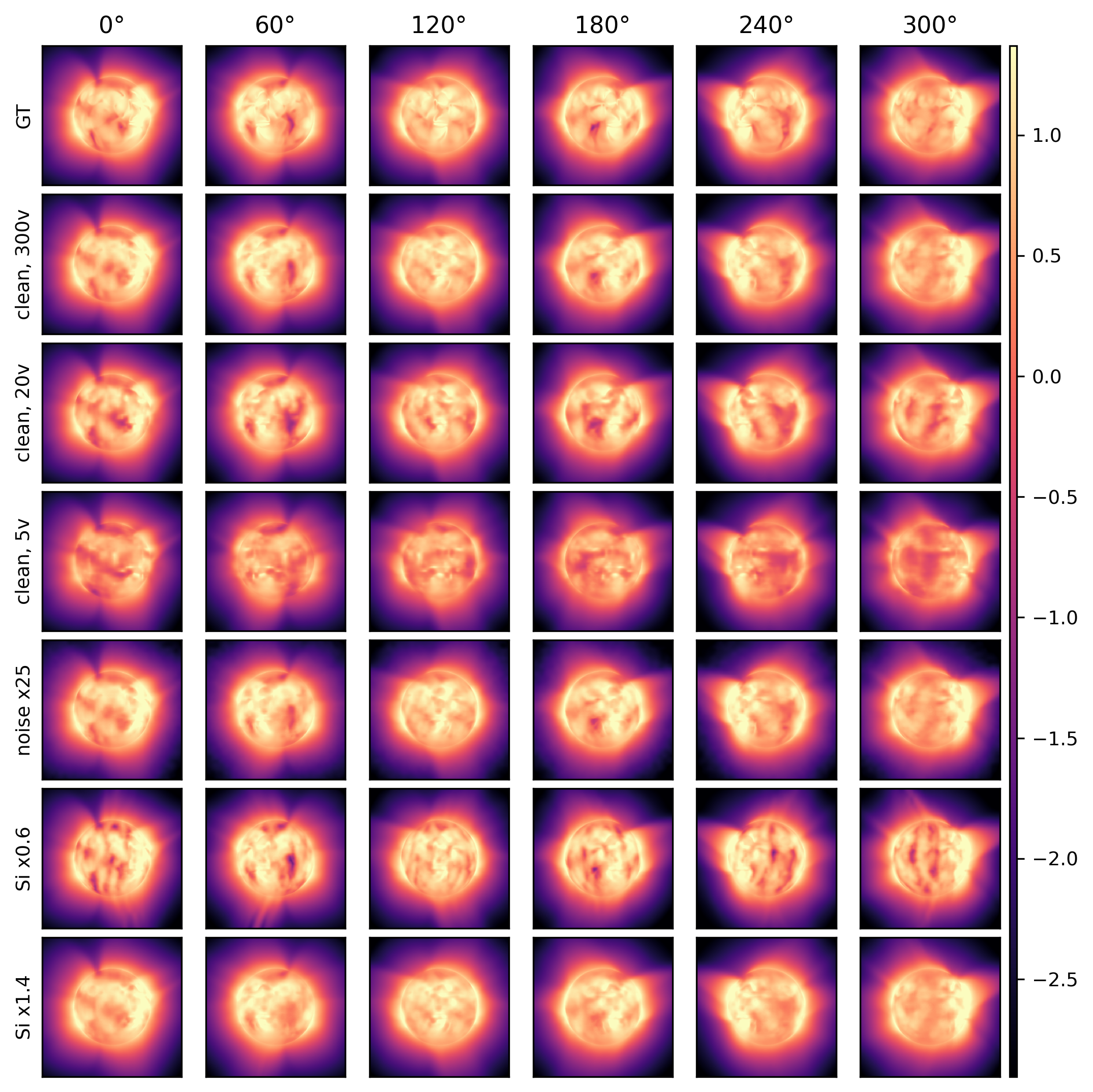}
	\end{center}
	\caption{LOS projections of total-emissivity log-intensities for selected conditions and view angles.}
	\label{fig:benchF_filmstrip_emissivity}
\end{figure}

This section provides renders of the column density in Figure \ref{fig:benchF_filmstrip_density}, total-emissivity weighted temperature in Figure \ref{fig:benchF_filmstrip_temp}, and total emissivity in Figure \ref{fig:benchF_filmstrip_emissivity}, for selected conditions and longitude view angles.

\end{document}